\documentclass[preprint,12pt]{elsarticle}
\usepackage{amssymb}
\usepackage{amsthm}
\usepackage{amsmath}
\allowdisplaybreaks[3]
\usepackage{amsfonts}
\usepackage{bm}
\usepackage{siunitx}
\usepackage{longtable}
\usepackage{algorithm}
\usepackage{algorithmic}
\usepackage{subcaption}
\usepackage{float}
\usepackage{placeins}
\usepackage{verbatim}

\journal{}

\begin{document}

\begin{frontmatter}

\title{Fixed Point Neural Acceleration and Inverse Surrogate Model for Battery Parameter Identification}

\author{Hojin Cheon\fnref{sgu}}
\ead{ghwlss@sogang.ac.kr}
\author{Hyeongseok Seo\fnref{sgu}}
\ead{fluorite@sogang.ac.kr}
\author{Jihun Jeon\fnref{sgu}}
\ead{wlgns97@sogang.ac.kr}
\author{Wooju Lee\fnref{hmc}}
\ead{wooju@hyundai.com}
\author{Dohyun Jeong\fnref{hmc}}
\ead{dohyun.jeong@hyundai.com}
\author{Hongseok Kim\corref{cor}\fnref{sgu}}
\ead{hongseok@sogang.ac.kr}

\affiliation[sgu]{organization={Department of Electronic Engineering, Sogang University}, 
            addressline={35 Baekbeom-ro}, 
            city={Seoul},
            postcode={04107}, 
            country={Republic of Korea}}
\affiliation[hmc]{organization={Hyundai Motor Company}, 
            addressline={93-20, Hyundaiyeonguso-ro}, 
            city={Hwaseong},
            postcode={18278}, 
            country={Republic of Korea}}

\begin{abstract}
The rapid expansion of electric vehicles has intensified the need for accurate and efficient diagnosis of lithium-ion batteries. Parameter identification of electrochemical battery models is widely recognized as a powerful method for battery health assessment. However, conventional metaheuristic approaches suffer from high computational cost and slow convergence, and recent machine learning methods are limited by their reliance on constant current data, which may not be available in practice. To overcome these challenges, we propose deep learning-based framework for parameter identification of electrochemical battery models. The proposed framework combines a neural surrogate model of the single particle model with electrolyte (NeuralSPMe) and a deep learning-based fixed-point iteration method. NeuralSPMe is trained on realistic EV load profiles to accurately predict lithium concentration dynamics under dynamic operating conditions while a parameter update network (PUNet) performs fixed-point iterative updates to significantly reduce both the evaluation time per sample and the overall number of iterations required for convergence. Experimental evaluations demonstrate that the proposed framework accelerates the parameter identification by more than 2000 times, achieves superior sample efficiency and more than 10 times higher accuracy compared to conventional metaheuristic algorithms, particularly under dynamic load scenarios encountered in practical applications.
\end{abstract}

\begin{keyword}
Lithium-ion Battery \sep
Deep Learning \sep
Parameter Identification \sep
Battery Model \sep
Battery Management System
\end{keyword}

\end{frontmatter}

\section{Introduction}
\label{Introduction}
With the rapid growth of electric vehicles (EVs), the diagnosis of lithium-ion batteries has become a critical issue. Battery diagnosis includes various tasks such as tracking capacity fade~\cite{choi2019machine,li2021online,figgener2024multi,tian2024exploiting} and increase in internal resistance~\cite{pozzato2023analysis,xu2024quantitative,pang2024accurate,pan2025identification}, forecasting remaining useful life~\cite{chen2024attmoe,pang2024accurate}, and detecting anomalies or potential faults~\cite{jeon2024proadd,kumar2025hybrid}. These capabilities support safety assurance and predictive maintenance of EV batteries. 
Among various diagnostic techniques, parameter identification of electrochemical battery models is recognized as a powerful method for assessing battery status~\cite{streb2023diagnosis,reza2024recent}, since each estimated parameter represents a specific physical characteristic and thus provides interpretable and reliable results. However, conventional approaches require significant computational effort, limiting their applicability in real-world EV battery analysis.

Traditionally, the parameter identification of the pseudo-two-dimensional (P2D) model~\cite{doyle1993modeling} or the single particle model with electrolyte (SPMe)~\cite{marquis2019asymptotic} has been performed by optimizing the battery parameters to minimize the error between the simulated and target voltage sequences, typically using metaheuristic algorithms such as genetic algorithms (GA)~\cite{zhang2014multi,jokar2016inverse} and particle swarm optimization (PSO)~\cite{rahman2016electrochemical,wimarshana2023multi,wang2023lithium}. However, the slow convergence of these algorithms has made the parameter identification challenging. For example, Li et al. compared 78 different metaheuristic algorithms for parameter identification of the P2D model and reported that teaching-learning-based optimization (TLBO) achieved the minimum error at the cost of 35,620 evaluations, whereas the cross-entropy method (CEM) achieved an acceptable error at 8,571 evaluations~\cite{li2024comparative}. 

Due to the high computational cost of metaheuristic algorithms, recent studies have attempted to incorporate deep learning techniques to enhance their efficiency. For instance, to accelerate the convergence of GA, Kim et al. integrated a neural network that predicts the parameters of the P2D model from a replay buffer generated by GA~\cite{kim2022parameter}, Xu et al. and Zhao et al. used a deep learning model to generate initial guesses of the parameters~\cite{xu2022enabling,zhao2024enhancing}. In another work, Wang et al. utilized a classifier to enhance the efficiency of PSO, by filtering particles which are not likely to converge~\cite{wang2024parameter}. These approaches greatly improved the efficiency of parameter identification, but still require a large number of evaluations. 
In parallel, deep learning methods that directly predict battery parameters also have been proposed to bypass metaheuristic algorithms. For example, Li et al. extracted 9 features from constant current (CC) charge and discharge data and trained a convolutional neural network (CNN) to predict the battery parameters~\cite{li2024deep}, while Ko et al. trained a transformer model on 50,000 simulated discharge curves~\cite{ko2024using}. Such approaches are much more efficient in terms of computational cost but have some limitations. One of these limitations is that the predicted parameters cannot be adjusted when the values are not accurate enough, or not feasible. Moreover, most of these methods have limited adaptability to practical load profiles, as they are trained on CC charge and discharge profiles. 

To address these issues, we propose a neural network named parameter update network (PUNet), which performs fixed-point iteration to update the battery model parameters. PUNet takes the reference voltage, current, evaluated voltage and the parameters as inputs, and outputs the updated parameters. In addition, the evaluated lithium concentration is also provided to PUNet, which significantly improves the accuracy of the updated parameters. PUNet shows superior sample efficiency compared to conventional optimization algorithms, as it leverages prior knowledge of the objective while conventional optimization methods do not. Specifically, the number of required evaluations is reduced by more than 24 times, compared to the covariance matrix adaptation evolution strategy (CMA-ES)~\cite{6790628}. For 10 parameter identification tasks, the average number of iterations of PUNet to make the root mean squared error (RMSE) of the terminal voltage less than 5 mV is only 16.3, while CMA-ES needs 402.6 iterations on average.

To further accelerate the parameter identification, the time required to evaluate the voltage and lithium concentration has to be reduced, and a surrogate model can be used. For example, Nascimento et al. proposed a hybrid model by combining reduced-order model and neural network to evaluate the terminal voltage~\cite{nascimento2021hybrid}. The reduced-order model has a low computation cost and can adapt to dynamic load profiles, but does not provide lithium concentration. Hassanaly et al. proposed a physics-informed neural network (PINN) surrogate model, which can adapt to dynamic loads and provide lithium concentration~\cite{hassanaly2024pinn}. However, the PINN surrogate model requires training for each load profile, which limits its evaluation speed. Zhang et al. proposed a parameter identification framework that utilizes a neural network-based surrogate model, but it is only applicable to \emph{constant} current data~\cite{zhang2025multi}. 

In this regard, we propose a physics-embedded neural surrogate model of the SPMe, named NeuralSPMe. The proposed model is designed to evaluate the lithium concentration and the terminal voltage under EV load profiles in a single forward pass. NeuralSPMe employs a nonlinear solver to determine the initial stoichiometries of the electrodes from the initial voltage and capacity-related parameters, followed by coulomb counting to calculate the average stoichiometries over time. Subsequently, the lithium concentrations at the electrode surfaces and the average concentrations in the electrolyte are predicted by a transformer encoder. The terminal voltage is then calculated using the predicted lithium concentrations, the battery model parameters, and the current. Due to its physics-embedded structure, NeuralSPMe successfully evaluates the terminal voltage with an average RMSE of 0.84 mV with respect to the SPMe, while that of data-driven model is substantially high, e.g., 10.23 mV. Moreover, the evaluation of the terminal voltage and the lithium concentration is significantly accelerated by using NeuralSPMe, achieving a speedup of 49.6$\times$ compared to the SPMe implemented in PyBaMM~\cite{sulzer2021python}. Consequently, the overall acceleration is more than 2100$\times$ when combined with PUNet.

The main contributions of this paper are as follows:
\begin{itemize}
    \item We propose NeuralSPMe, a physics-embedded neural surrogate model of the SPMe. Instead of directly predicting terminal voltage, NeuralSPMe predicts lithium concentration and leverages the voltage expression to capture the electrochemical dynamics of batteries more efficiently. To minimize the complexity of lithium concentration prediction, a simple nonlinear solver and coulomb counting are integrated into a transformer encoder architecture.
    \item We propose PUNet and a fixed-point iteration method for fast and accurate parameter identification. PUNet updates the parameters using the reference voltage, current, evaluated voltage, lithium concentration, and the parameters themselves. PUNet is trained with the parameter labels that conventional optimization algorithms cannot access, improving both the number of updates and accuracy. The evaluated voltage and lithium concentration can be provided by NeuralSPMe for faster evaluation.
    \item NeuralSPMe speeds up parameter identification by 49.6 times compared to the SPMe and also achieves the RMSE of only \SI{0.84}{\milli\volt} under EV load profiles. At the same time, PUNet reduces the number of iterations by 24.7 times and achieved 10 times higher accuracy compared to CMA-ES. Consequently, the proposed framework takes only 1.32 seconds for parameter identification, which is more than 2100 times acceleration compared to the conventional approach using CMA-ES and SPMe.
\end{itemize}

\section{Methodology}
\label{Methodology}

\subsection{Problem Formulation}
\label{Problem Formulation}
One typical way of battery parameter identification under dynamic operating conditions is to minimize the error between reference and simulated voltage sequences. Specifically, battery parameter identification can be formulated as an optimization problem over $\bm{\lambda}$ (the battery model parameters) as follows:
\begin{equation}
    \label{eq:objective}
    \min_{\bm{\lambda}} \sum_{i=1}^{M} \lvert V_{i}^{\text{ref}} - V_{i}^{\text{sim}}(\bm{\lambda},I_{i}) \rvert,
\end{equation}
where $V_{i}^{\text{ref}}$ is the $i$-th reference voltage \emph{sequence}, $M$ is the number of sequences, $V_{i}^{\text{sim}}(\bm{\lambda},I_{i})$ is the output of the battery simulator, $I_{i}$ is a current sequence. This approach is valid when the electrochemical battery simulator is highly accurate, but searching over feasible $\bm{\lambda}$ is not trivial considering the slow speed of the battery simulator. To overcome this, we propose a neural surrogate model of SPMe, called NeuralSPMe, for fast evaluation of $V_{i}^{\text{sim}}(\bm{\lambda},I_{i})$. Let $V$ denote a voltage sequence predicted by NeuralSPMe. For notational simplicity, we omit the sequence index $i$ unless it is specifically needed. Then, the voltage errors made by $V^{\text{sim}}$ and $V$ with respect to $V^{\text{ref}}$ can be expressed using the triangle inequality such as
\begin{equation}
    \label{eq:objective1}
    \lvert V^{\text{ref}} - V^{\text{sim}} \rvert \leq \lvert V^{\text{ref}} - V \rvert + \lvert V^{\text{sim}} - V \rvert.
\end{equation}
Note that $V^{\text{sim}}$ and $V$ are functions of the parameters $\bm{\lambda}$ and the current $I$.

Equation~(\ref{eq:objective1}) indicates that the NeuralSPMe can be used for parameter identification when the error between the simulated voltage ($V^{\text{sim}}$) and the neural predicted voltage ($V$) is small. Hence, leveraging $V$ obtained from the NeuralSPMe instead of $V^{\text{sim}}$ can serve as an upper bound of~(\ref{eq:objective}). In this regard, the proposed framework aims to identify the parameters $\bm{\lambda}$ that minimize $\lvert V^{\text{ref}} - V \rvert$ as well as $\lvert V^{\text{sim}} - V \rvert$ in~(\ref{eq:objective1}).

\subsection{Proposed Framework}
\label{Proposed Framework}
We consider the following two mappings: $\bm{\Phi}_{\text{v}}$ mapping from the parameters-current pairs to the voltage, 
\begin{equation}
\label{eq:mapping_Phi}    
V = \bm{\Phi}_{\text{v}}(\bm{\lambda}, I),
\end{equation}
and $\bm{\Psi}$ mapping from the voltage-current pairs to the parameters,
\begin{equation}
\label{eq:mapping_Psi}
\bm{\lambda} = \bm{\Psi}(V, I).
\end{equation}
As expressed in~(\ref{eq:mapping_Phi}), while the first mapping $\bm{\Phi}_{\text{v}}$ is a battery model that calculates the terminal voltage given the parameters $\bm{\lambda}\in\mathbb{R}^{d}$ and the current $I\in\mathbb{R}^{T}$, the second mapping $\bm{\Psi}$ estimates the parameters $\bm{\lambda}$ given the terminal voltage $V\in\mathbb{R}^{T}$ and the current $I$.
Once the parameters are correctly identified, denoted by $\bm{\lambda}^{*}$, the terminal voltage calculated from the identified parameters approximates the reference voltage. In that case, the identified parameters $\bm{\lambda}^{*}$ is a fixed-point such as:
\begin{equation}
    \label{eq:fixed_point}
    \bm{\lambda}^{*}=\bm{\Psi}(\bm{\Phi}_{\text{v}}(\bm{\lambda}^{*},I), I).
\end{equation}
Starting from this key idea, we further refine $\bm{\Phi}_{\text{v}}$ and $\bm{\Psi}$ with more inputs as follows. Specifically, at the $k$-th iteration in finding the fixed-point of~(\ref{eq:fixed_point}), the parameter estimator $\bm{\Psi}$ can further utilize additional information such as the intermediate parameters itself ($\bm{\lambda}^{k}$), the evaluated terminal voltage ($V^{k}$), and the corresponding lithium concentration ($\textbf{c}^{k}$) to improve the accuracy of parameter update:
\begin{align}
    \bm{\lambda}^{k+1} &= \bm{\Psi}(V^{k},\textbf{c}^{k},\bm{\lambda}^{k},V^{\text{ref}},I) = \bm{\Psi}(\textbf{U}^{k}), \label{eq:mappings2}\\
    \textbf{U}^{k} &\triangleq [V^{k}, \textbf{c}^{k}, \bm{\lambda}^{k}, V^{\text{ref}}, I]. \label{eq:U_k}
\end{align}
In addition, the battery model $\bm{\Phi}_{\text{v}}$ is modified to $\bm{\Phi}_{\text{c}}$ in order to evaluate the lithium concentration $\textbf{c}^{\text{k}}$ as in~(\ref{eq:nspme_simple}) 
\begin{equation}
    \textbf{c}^{k} = \bm{\Phi}_{\text{c}}(\bm{\lambda}^{k},I), \label{eq:nspme_simple}
\end{equation}
and then the terminal voltage $V^{k}$ is calculated from the voltage expression of SPMe as expressed in~(\ref{eq:c_to_v}) using a function $h$,
\begin{equation}
V^{k} = h(\textbf{c}^{k}, \bm{\lambda}^{k}, I). \label{eq:c_to_v}
\end{equation}
The advantage of using~(\ref{eq:nspme_simple}) and~(\ref{eq:c_to_v}) instead of~(\ref{eq:mapping_Phi}) lies in that it reduces the burden of the neural network and thus enables fast and accurate learning. Notations used in this explanation are summarized in Table~\ref{table:Notation_deep learning}.
\begin{table}[h]
    \centering
    \begin{tabular}{c p{0.8\linewidth}}
        \hline
        \textbf{Symbol} & \textbf{Description} \\
        \hline
        $I$ & Applied Current \\
        $\mathring{\bm{\lambda}}$ & Battery model parameter label \\
        $\bm{\lambda}$ & Output parameters of PUNet \\
        $\bm{\lambda}^{k}$ & Parameters at k-th iteration \\
        $\bm{\lambda}^{*}$ & Fixed-point of the parameters (Identified parameters) \\
        $\tilde{\bm{\lambda}}$ & Perturbed parameters \\
        $\textbf{x}$ & Input sequence of NeuralSPMe with actual parameters\\
        $\textbf{x}^{k}$ & Input sequence of NeuralSPMe at k-th iteration \\
        $\textbf{x}^{*}$ & Input sequence of NeuralSPMe with identified parameters \\
        $\tilde{\textbf{x}}^{*}$ & Input sequence of NeuralSPMe with perturbed parameters \\
        $\textbf{\r{y}}$ & Normalized lithium concentration label \\
        $\textbf{y}$ & Output sequence of NeuralSPMe with actual parameters\\
        $\textbf{y}^{k}$ & Output sequence of NeuralSPMe at k-th iteration \\
        $\textbf{y}^{*}$ & Output sequence of NeuralSPMe with identified parameters \\
        $\tilde{\textbf{y}}$ & Output sequence of NeuralSPMe with perturbed parameters \\
        $V^{\text{ref}}$ & Reference voltage \\
        $V^{\text{sim}}$ & Voltage simulated with SPMe \\
        $V^{k}$ & Voltage predicted with parameters at k-th iteration \\
        $\tilde{V}$ & Voltage predicted with perturbed parameters\\
        $\bm{\Phi}$ & Neural surrogate model of SPMe (NeuralSPMe)\\
        $\bm{\Psi}$ & Parameter estimator network (PUNet) \\
        \hline
    \end{tabular}
    \caption{Notations summary.}
    \label{table:Notation_deep learning}
\end{table}

\begin{figure}[H]
    \begin{center}
        \includegraphics[width=\textwidth]{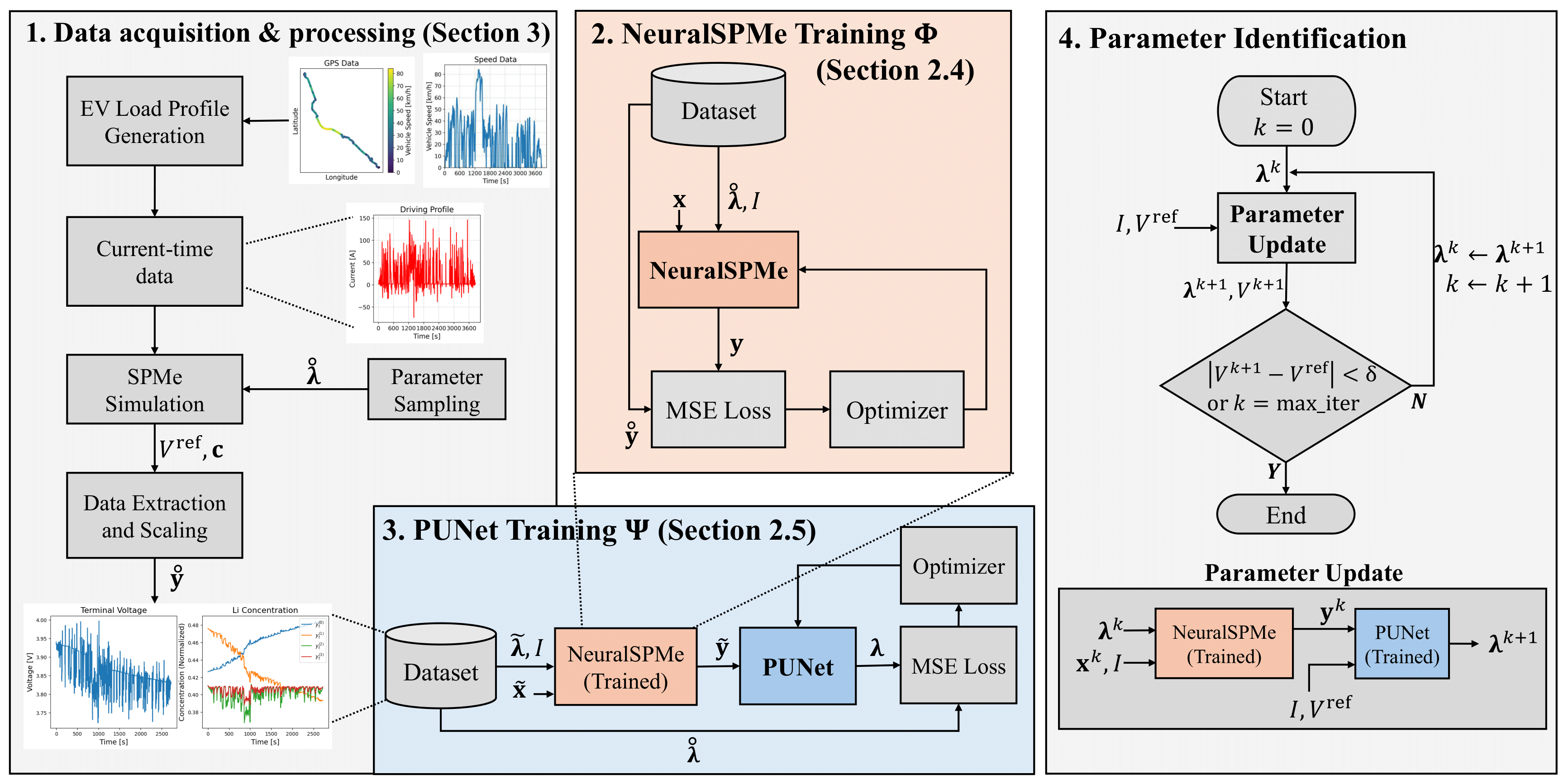}
        \caption{The proposed framework.}
        \label{fig:Framework}
    \end{center}
\end{figure}

The proposed framework is illustrated in Figure~\ref{fig:Framework}. The framework consists of three main steps: NeuralSPMe training, PUNet training, and parameter identification. The architectures of the proposed NeuralSPMe and PUNet will be described in Section~\ref{NeuralSPMe} and Section~\ref{PUNet} after discussing $h(\textbf{c}^{k}, \bm{\lambda}^{k}, I)$ in~(\ref{eq:c_to_v}), i.e., the voltage expression of SPMe in Section~\ref{SPMe}.
Data acquisition of making a set of labeled parameters $\mathring{\bm{\lambda}}$ and preprocessing will be given in Section~\ref{Data Acquisition and Preprocessing} as a part of experimental setup.

\subsection{Single Particle Model with Electrolyte (SPMe)}
\label{SPMe}
We employ SPMe~\cite{marquis2019asymptotic} as an electrochemical model of $V^{\text{sim}}$. Note that SPMe is a simplified version of the P2D model that has expensive differential algebraic equations~\cite{doyle1993modeling}. The relatively low computational cost of the SPMe supports the construction of a large dataset for training the neural networks while maintaining acceptable precision. Moreover, the terminal voltage can be calculated without integration so that the NeuralSPMe can have a compact architecture. The full version of voltage expression $h$ in~(\ref{eq:c_to_v}) for the SPMe is shown in~(\ref{eq:V_expression}), with the associated symbols and their descriptions are summarized in Table~\ref{table:Notation_SPMe}, unless specified otherwise;
\begin{table}[H]
    \centering
    \begin{tabular}{c p{0.7\linewidth}}
        \hline
        \textbf{Symbol} & \textbf{Description} \\
        \hline
        $J$ & Current density \\
        $Q_{\text{Li}}$ & Cyclable lithium capacity \\
        $\text{k} \in \text{\{p,sep,n\}}$ & Domain index \\
        $c_{\text{s,k}}$ & Lithium concentration in solid phase \\
        $c_{\text{s,k}}^{\text{surf}}$ & Lithium concentration in electrode particle surface \\
        $c_{\text{s,k}}^{\text{max}}$ & Maximum lithium concentration in solid phase \\
        $\theta_{\text{k}}$ & Stoichiometry of the electrode \\
        $\theta_{\text{k}}^{0}, \theta_{\text{k}}^{100}$ & Stoichiometry of the electrode at 0\%, 100\% SoC \\
        $c_{\text{e,k}}$ & Lithium concentration in electrolyte phase \\
        $\overline{c_{\text{e,k}}}$ & Average lithium concentration in electrolyte phase \\
        $c_{\text{e,typ}}$ & Typical lithium concentration in electrolyte phase \\
        $D_{\text{s,k}}$ & Solid phase diffusion coefficient \\
        $D_{\text{e}}$ & Electrolyte diffusion coefficient \\
        $L_{\text{k}}$ & Thickness of the electrode \\
        $\varepsilon_{\text{k}}$ & Porosity of the electrode \\
        $\sigma_{\text{k}}$ & Solid phase conductivity \\
        $a_{\text{k}}$ & Specific interfacial area \\
        $j_{0,\text{k}}$ & Exchange-current density \\
        $m_{\text{k}}$ & Reaction rate constant \\
        $E_{\text{a,k}}$ & Activation energy \\
        $\kappa_{\text{e}}$ & Electrolyte conductivity \\
        $\beta$ & Bruggeman coefficient \\
        $t^{+}$ & Transference number \\
        $U_{\text{k}}$ & Open circuit potential of the electrode \\
        $\eta_{\text{r}}$ & Reaction overpotential \\
        $\eta_{\text{c}}$ & Concentration overpotential \\
        $\Delta\Phi_{\text{Elec}}$ & Electrolyte phase Ohmic drop \\
        $\Delta\Phi_{\text{Solid}}$ & Solid phase Ohmic drop \\
        $N$ & Number of electrodes \\
        $A$ & Electrode surface area \\
        $F$ & Faraday constant \\
        $R$ & Gas constant \\
        $\mathcal{T}$ & Temperature \\
        \hline
    \end{tabular}
    \caption{Physical quantities and parameters in the SPMe formulation.}
    \label{table:Notation_SPMe}
\end{table}

\begin{subequations}
    \label{eq:V_expression}
    \begin{align}
        V &= U_{\text{eq}} + \eta_{\text{r}} + \eta_{\text{c}} + 
        \Delta\Phi_{\text{Elec}} + \Delta\Phi_{\text{Solid}}, \\
        U_{\text{eq}} &= U_{\text{p}}(c_{\text{s,p}}^{\text{surf}}) -
        U_{\text{n}}(c_{\text{s,n}}^{\text{surf}}), \\
        \eta_{\text{r}} &= \frac{2RT}{F} \sinh^{-1}\left( \frac{-J}{2a_{\text{p}}j_{0,\text{p}}L_{\text{p}}} \right)\\
        &- \frac{2RT}{F} \sinh^{-1}\left( \frac{J}{2a_{\text{n}}j_{0,\text{n}}L_{\text{n}}} \right), \nonumber\\
        j_{\text{0,p}} &= \frac{1}{L_{\text{p}}}m_{\text{p}}\exp\left(\frac{E_{\text{a,p}}}{R}\left(\frac{1}{\mathcal{T}_{\text{ref}}}-\frac{1}{\mathcal{T}}\right)\right)\\
        & \times (c_{\text{s,p}}^{\text{surf}})^{0.5}(c_{\text{s,p}}^{\text{max}}-c_{\text{s,p}}^{\text{surf}})^{0.5}(\overline{c_{\text{e,p}}^{0.5}}), \nonumber\\
        j_{\text{0,n}} &= \frac{1}{L_{\text{n}}}m_{\text{n}}\exp\left(\frac{E_{\text{a,n}}}{R}\left(\frac{1}{\mathcal{T}_{\text{ref}}}-\frac{1}{\mathcal{T}}\right)\right)\\
        & \times (c_{\text{s,n}}^{\text{surf}})^{0.5}(c_{\text{s,n}}^{\text{max}}-c_{\text{s,n}}^{\text{surf}})^{0.5}(\overline{c_{\text{e,n}}^{0.5}}), \nonumber\\
        \overline{c_{\text{e,p}}^{0.5}} &= \int_{L-L_{\text{p}}}^{L} (c_{\text{e,p}})^{0.5} dx, \quad
        \overline{c_{\text{e,n}}^{0.5}} = \int_{0}^{L_{\text{n}}} (c_{\text{e,n}})^{0.5} dx, \nonumber\\
        \eta_{\text{c}} &= \frac{2RT}{Fc_{\text{e,typ}}}(1-t^{+})(\overline{c_{\text{e,p}}}-\overline{c_{\text{e,n}}}), \\\
        \overline{c_{\text{e,p}}} &= \int_{L-L_{\text{p}}}^{L} c_{\text{e,p}} dx, \quad
        \overline{c_{\text{e,n}}} = \int_{0}^{L_{\text{n}}} c_{\text{e,n}} dx, \nonumber \\
        \Delta\Phi_{\text{Elec}} &= \frac{J}{\kappa_{\text{e}}(c_{\text{e,typ}})}\left(\frac{L_{\text{n}}}{3\varepsilon_{\text{n}}^{\beta}}+\frac{L_{\text{sep}}}{\varepsilon_{\text{sep}}^{\beta}}+\frac{L_{\text{p}}}{3\varepsilon_{\text{p}}^{\beta}}\right), \\
        \Delta\Phi_{\text{Solid}} &= -\frac{J}{3}
        \left( \frac{L_{\text{p}}}{\sigma_{\text{p}}} + 
        \frac{L_{\text{n}}}{\sigma_{\text{n}}} \right), \\
        J &= \frac{I}{NA}.
    \end{align}
\end{subequations}

Following the voltage expression in~(\ref{eq:V_expression}), the terminal voltage $V$ can be calculated with 6 lithium concentration values, associated parameters, and the current density. The lithium concentration at the solid surface ($c_{\text{s,p}}^{\text{surf}}$, $c_{\text{s,n}}^{\text{surf}}$) are required for the open-circuit potential ($U_{\text{p}}$, $U_{\text{n}}$) and the exchange-current density ($j_{\text{0,p}}$, $j_{\text{0,n}}$), while the average lithium concentration in each side of electrolyte ($\overline{c_{\text{e,p}}}$, $\overline{c_{\text{e,n}}}$) is needed for the concentration overpotential ($\eta_{\text{c}}$). In addition, the average of the square root of the lithium concentration in each side of electrolyte ($\overline{c_{\text{e,p}}^{0.5}}$, $\overline{c_{\text{e,n}}^{0.5}}$) is needed for the exchange-current density. In the proposed framework, these 6 lithium concentration values are predicted by NeuralSPMe. A comprehensive explanation of the SPMe can be found in~\cite{marquis2019asymptotic}.

\subsection{NeuralSPMe}
\label{NeuralSPMe}

\subsubsection{Physics-Embedded Output Structure}
As presented in~(\ref{eq:nspme_simple}), NeuralSPMe $\bm{\Phi}_{\text{c}}$ evaluates the lithium concentration from the current and the parameters, in place of SPMe. By evaluating the lithium concentration instead of directly predicting the terminal voltage, PUNet can leverage the concentration to update the parameters. Furthermore, predicting the concentration allows for a physics-embedded design of the NeuralSPMe, which makes it more accurate and efficient to train. 

At each time step, NeuralSPMe evaluates 6 lithium concentration values: $c_{\text{s,p}}^{\text{surf}}$, $c_{\text{s,n}}^{\text{surf}}$, $\overline{c_{\text{e,p}}}$, $\overline{c_{\text{e,n}}}$, $\overline{c_{\text{e,p}}^{0.5}}$, and $\overline{c_{\text{e,n}}^{0.5}}$.
However, as lithium in the electrolyte is conserved in the SPMe, it is not necessary to predict the average lithium concentration in the electrolyte phase for the both sides. 
We slightly modify $\bm{\Phi}_{\text{c}}$ from~(\ref{eq:nspme_simple}) to have the final NeuralSPMe $\bm{\Phi}$ in~(\ref{eq:NSPMe_output_a}) so that it takes an additional input sequence $\textbf{x}\in\mathbb{R}^{4\times T}$ other than $\bm{\lambda}$ and $I$ and generates an output sequence $\textbf{y}\in\mathbb{R}^{4\times T}$, which is a \emph{normalized} representation of the concentration $\textbf{c}\in\mathbb{R}^{6\times T}$ as shown in~(\ref{eq:NSPMe_output_conc})
\begin{equation}
\textbf{y}=\bm{\Phi}\left(\textbf{x},\bm{\lambda},I\right). \label{eq:NSPMe_output_a}
\end{equation}
The actual concentration values at time step $t$ are computed from the output sequence $\textbf{y}$ at time step $t$, i.e., $\textbf{y}(t)=\left[y_{0}(t),y_{1}(t),y_{2}(t),y_{3}(t)\right]^{\top}$ as follows:
\begin{subequations}
\label{eq:NSPMe_output_conc}
\begin{align}
    c_{\text{p}}^{\text{surf}}\left(t\right)&=c_{\text{s,p}}^{\text{max}}y_{0}(t), \\
    c_{\text{n}}^{\text{surf}}\left(t\right)&=c_{\text{s,n}}^{\text{max}}y_{1}(t), \\
    \overline{c_{\text{e,p}}}\left(t\right)&=\frac{L_{\text{p}}\varepsilon_{\text{p}}+L_{\text{n}}\varepsilon_{\text{n}}}{L_{\text{p}}\varepsilon_{\text{p}}}c_{\text{e,typ}}y_{2}(t), \\
    \overline{c_{\text{e,n}}}\left(t\right)&=\frac{L_{\text{p}}\varepsilon_{\text{p}}+L_{\text{n}}\varepsilon_{\text{n}}}{L_{\text{n}}\varepsilon_{\text{n}}}c_{\text{e,typ}}\left(1-y_{2}(t)\right), \\
    \overline{c_{\text{e,p}}^{0.5}}\left(t\right)&=\frac{L_{\text{p}}\varepsilon_{\text{p}}+L_{\text{n}}\varepsilon_{\text{n}}}{L_{\text{p}}\varepsilon_{\text{p}}}c_{\text{e,typ}}^{0.5}y_{3}(t), \\
    \overline{c_{\text{e,n}}^{0.5}}\left(t\right)&=\frac{L_{\text{p}}\varepsilon_{\text{p}}+L_{\text{n}}\varepsilon_{\text{n}}}{L_{\text{n}}\varepsilon_{\text{n}}}c_{\text{e,typ}}^{0.5}\left(1-y_{3}(t)\right).
\end{align}
\end{subequations}
Or, simply 
\begin{equation}
    \textbf{c}(t)=\text{H}\textbf{y}(t)
\end{equation}
where $\text{H}\in\mathbb{R}^{6\times 4}$ that determines the linear relationship.
Here, $y_{0}(t)$ and $y_{1}(t)$ represent the surface stoichiometries of the positive and negative particles, while $y_{2}(t)$ and $y_{3}(t)$ represent the fraction of lithium in the electrolyte phase on the positive side, and the fraction of square root of that, respectively. 

\subsubsection{Physics-Embedded Input Structure}
To evaluate the surface stoichiometries, NeuralSPMe utilizes the \emph{average} stoichiometries of the each particles as initial estimates, which can easily be calculated from the initial stoichiometries and the current. 
Similarly, volume fraction of electrolyte in the positive side can be used as the initial estimates for $y_{2}(t)$ and $y_{3}(t)$. These initial estimates, denoted as $\textbf{x}$, are defined as follows:
\begin{subequations}
\label{eq:NSPMe_input}
\begin{align}
    \textbf{x}(t)&=\left[x_{0}(t),x_{1}(t),x_{2}(t),x_{3}(t)\right]^{\top}, \\
    x_{0}(t)&=x_{0}(0)+\frac{\sum_{\tau=0}^{t}I\left(\tau\right)\Delta\tau}{Q_{\text{p}}}, \\
    x_{1}(t)&=x_{1}(0)-\frac{\sum_{\tau=0}^{t}I\left(\tau\right)\Delta\tau}{Q_{\text{n}}}, \label{eq:NSPMe_input_d}\\
    x_{2}(0)&=x_{3}(0)=\frac{L_{\text{p}}\varepsilon_{\text{p}}}{L_{\text{p}}\varepsilon_{\text{p}}+L_{\text{n}}\varepsilon_{\text{n}}}, \\
    x_{2}(t)&=x_{2}(0),\quad x_{3}(t)=x_{3}(0),
\end{align}
\end{subequations}
where $Q_{\text{p}}=NAFL_{\text{p}}(1-\varepsilon_{\text{p}})c_{\text{s,p}}^{\text{max}}$ and $Q_{\text{n}}=NAFL_{\text{n}}(1-\varepsilon_{\text{n}})c_{\text{s,n}}^{\text{max}}$ are the capacities of the positive and negative electrodes, respectively. Please, refer to Table~\ref{table:Notation_SPMe} for the constants, e.g., $N$, $A$, $F$, etc.

In~(\ref{eq:NSPMe_input}), $x_{0}(0)$ and $x_{1}(0)$ denote the initial stoichiometries of the positive and negative particles. Initial stoichiometries are obtained by solving following equations for the 6 unknowns, $x_{0}(0)$, $x_{1}(0)$, $\theta_{\text{p}}^{100}$, $\theta_{\text{p}}^{0}$, $\theta_{\text{n}}^{100}$, and $\theta_{\text{n}}^{0}$, via the Newton-Raphson method,
\begin{subequations}
\label{eq:init_sto}
\begin{align}
    U_{\text{p}}(c_{\text{s,p}}^{\text{max}}\theta_{\text{p}}^{100}) - U_{\text{n}}(c_{\text{s,n}}^{\text{max}}\theta_{\text{n}}^{100}) = V^{100}, &\\
    U_{\text{p}}(c_{\text{s,p}}^{\text{max}}\theta_{\text{p}}^{0}) - U_{\text{n}}(c_{\text{s,n}}^{\text{max}}\theta_{\text{n}}^{0}) = V^{0}, &\\
    Q_{\text{p}}(\theta_{\text{p}}^{0}-\theta_{\text{p}}^{100}) = Q_{\text{n}}(\theta_{\text{n}}^{100}-\theta_{\text{n}}^{0}), &\\
    \left(Q_{\text{p}}\theta_{\text{p}}^{0}+Q_{\text{n}}\theta_{\text{n}}^{0}\right)/3600 = Q_{\text{Li}}, &\\
    U_{\text{p}}(c_{\text{s,p}}^{\text{max}}x_{0}(0))-U_{\text{n}}(c_{\text{s,n}}^{\text{max}}x_{1}(0)) = V^{\text{ref}}(0), &\\
    \frac{x_{0}(0)-\theta_{\text{p}}^{\text{0}}}{\theta_{\text{p}}^{100}-\theta_{\text{p}}^{0}} = \frac{x_{1}(0)-\theta_{\text{n}}^{\text{0}}}{\theta_{\text{n}}^{100}-\theta_{\text{n}}^{0}}&
\end{align}
\end{subequations}
where $V^{100}$ and $V^{0}$ are the terminal voltages at 100\% and 0\% state of charge (SoC), respectively.

\subsubsection{Network Architecture and Training}
\begin{figure}[t]
\begin{center}
    \includegraphics[width=\textwidth]{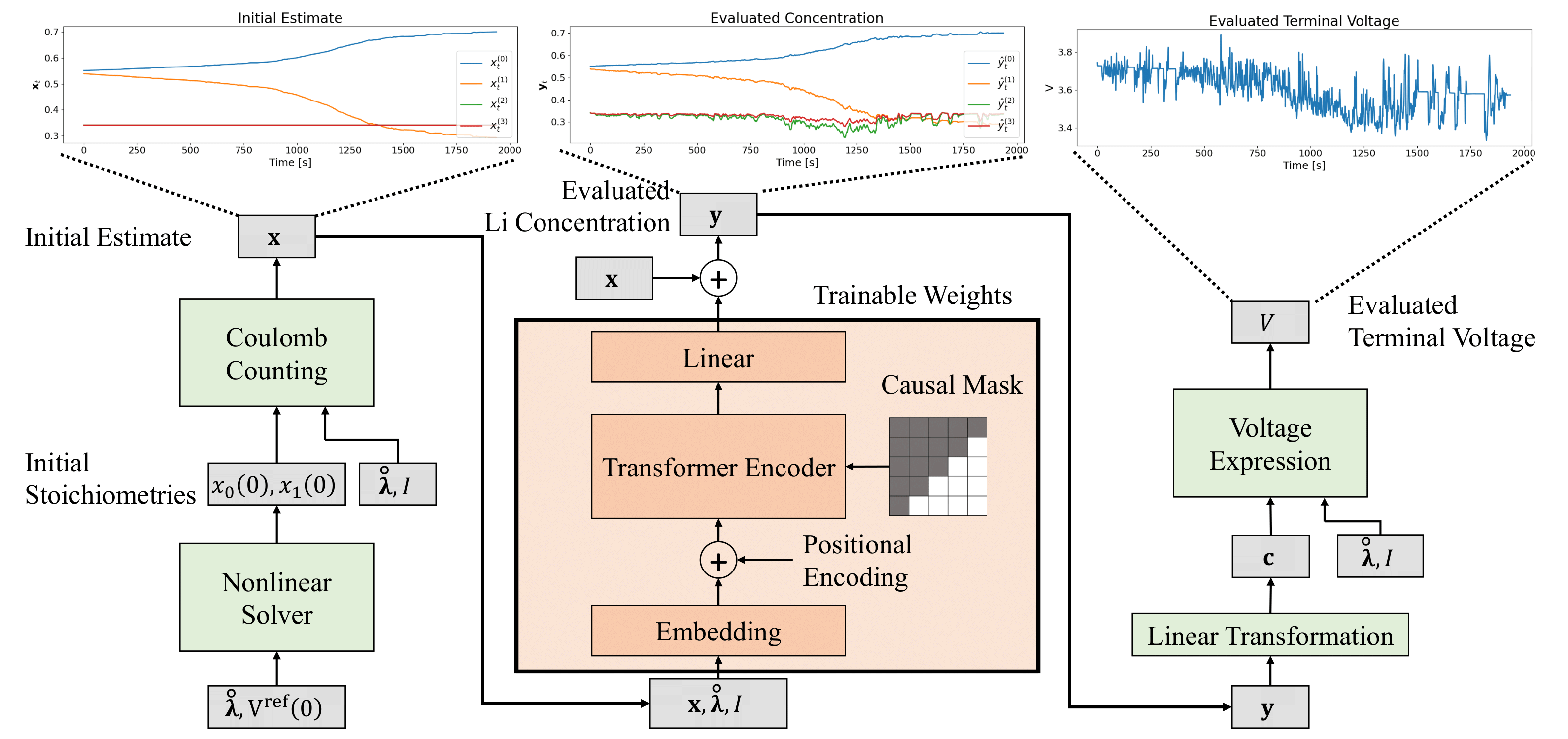}
    \caption{The architecture of NeuralSPMe $\bm{\Phi}$.}
    \label{fig:NSPMe}
\end{center}
\end{figure}

As shown in Figure~\ref{fig:NSPMe}, the proposed NeuralSPMe is based on a transformer encoder~\cite{vaswani2017attention}, which is well-known for its capability in handling long sequences. 
The parameters $\bm{\lambda}$ are repeated for each time step and concatenated with the input sequence $\textbf{x}$ and the current $I$, and provided to embedding layers. The embedding layers are two feed-forward layers. The sinusoidal positional encoding~\cite{vaswani2017attention} is applied to the embedded input, followed by a transformer encoder. The causal mask is applied to the transformer encoder to force the model to use only past time steps while evaluating the normalized lithium concentration $\textbf{y}$. The output of the transformer encoder is projected on 4-dimensional space using a feed-forward layer. Our loss function for NeuralSPMe $\bm{\Phi}$ is defined as a mean squared error over the output sequence:
\begin{equation}
    \label{eq:loss_nspme}
    L_{\Phi}=\sum_{\left(\mathring{\bm{\lambda}},I,\textbf{\r{y}}\right)\sim\mathcal{D}}\big|\big|\textbf{\r{y}}-\bm{\Phi}\left(\textbf{x},\mathring{\bm{\lambda}},I\right)\big|\big|^{2}_{\text{F}}
\end{equation}
where $\textbf{\r{y}}$ is the label generated from SPMe using the label parameter $\mathring{\bm{\lambda}}$ in a dataset $\mathcal{D}$, and $||\cdot||$ is the Frobenius norm.

\subsection{Parameter Update Network (PUNet)}
\label{PUNet}
As expressed in~(\ref{eq:mappings2}), PUNet is designed to update the parameters based on the target voltage, current, the evaluated voltage and concentration, and the parameters used for the evaluation.
The PUNet expressed in~(\ref{eq:PUNet_update}) is a reformulated version of~(\ref{eq:mappings2}) for multiple target voltage and current sequences, and $||$ denotes the concatenation operation.
\begin{equation}
\label{eq:PUNet_update}
\bm{\lambda}^{k+1}=\bm{\Psi}(\textbf{U}_{1}^{k} || \cdots || \textbf{U}_{M}^{k}),
\end{equation}
where 
\begin{align}
\textbf{U}_{i}^{k} &= \left[V_{i}^{k}, \textbf{y}_{i}^{k}, \bm{\lambda}^{k}, V_{i}^{\text{ref}}, I_{i}\right], \\
V_{i}^{k}&=h(\text{H}\textbf{y}^{k}_{i},\bm{\lambda}^{k}, I_{i}), \label{eq:V_i^k}\\
\textbf{y}^{k}_{i}&=\bm{\Phi}(\textbf{x}^{k}_{i},\bm{\lambda}^{k},I_{i}), \quad i=1,\dots, M.
\end{align}
Note that $\textbf{U}_{i}^{k}$ is the corresponding input for the $i$-th target voltage and current sequences at $k$-th iteration.

\begin{figure}[H]
    \begin{center}
        \includegraphics[width=0.7\columnwidth]{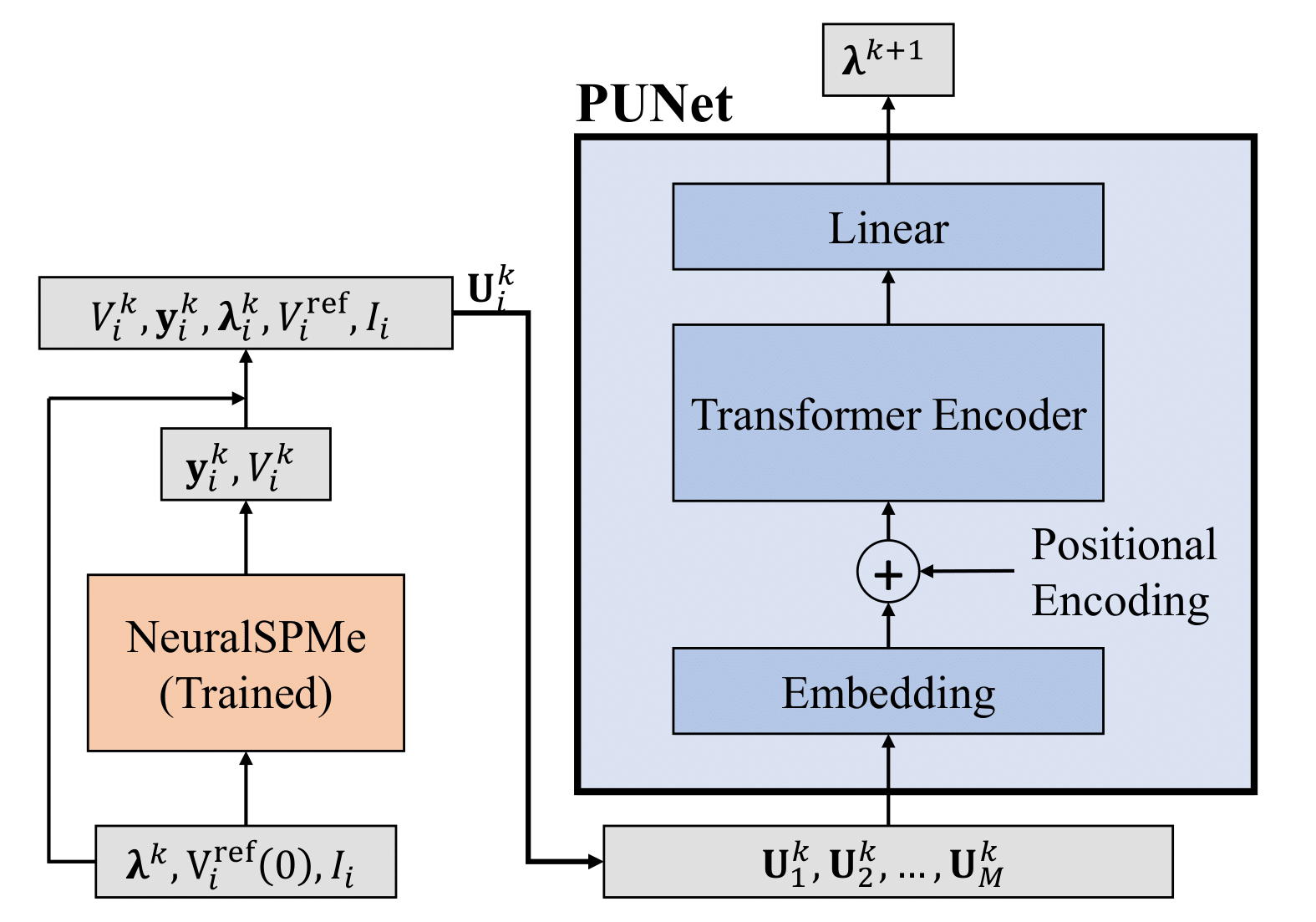}
        \caption{The architecture of PUNet $\bm{\Psi}$.}
        \label{fig:ParameterEstimator}
    \end{center}
\end{figure}

If the actual parameter is a fixed-point of a contraction mapping, our iterative approach converges. The loss function for PUNet in (\ref{eq:loss_punet}) is designed to meet these conditions.
\begin{equation}
\label{eq:loss_punet}
L_{\Psi}=\sum_{(V^{\text{ref}},I,\mathring{\bm{\lambda}}) \sim \mathcal{D}}\left(\left|\mathring{\bm{\lambda}}-\bm{\Psi}\left(\mathop{\Vert}\limits_{i=1}^{M}\tilde{\textbf{U}}_{i}\right)\right|^{2}+\left|\mathring{\bm{\lambda}}-\bm{\Psi}\left(\mathop{\Vert}\limits_{i=1}^{M}\textbf{U}_{i}\right)\right|^{2}\right),
\end{equation}
where
\begin{align}
\tilde{\textbf{U}}_{i}&=[\tilde{V}_{i},\tilde{\textbf{y}}_{i},\tilde{\bm{\lambda}},V^{\text{ref}}_{i},I_{i}], \label{eq:perturbed_U} \\
\tilde{V}_{i}&=h(\text{H}\tilde{\textbf{y}}_{i}, \tilde{\bm{\lambda}}, I_{i}), \\
\tilde{\textbf{y}}_{i}&=\bm{\Phi}(\tilde{\textbf{x}}_{i},\tilde{\bm{\lambda}},I_{i}),  \quad i=1,\dots , M \\
\tilde{\bm{\lambda}}&=\bm{\lambda}+\mathcal{N}(0,\sigma^{2}). \label{eq:perturbed_params}
\end{align}
In~(\ref{eq:perturbed_params}), $\tilde{\bm{\lambda}}$ denotes the perturbed parameters where $\sigma^{2}$ is the variance of Gaussian distribution. Using $\tilde{\bm{\lambda}}$, we make a perturbed input $\tilde{\text{U}}_{i}$ in~(\ref{eq:perturbed_U}), which serves to make PUNet a contraction mapping. The first term, named the \textit{contraction loss}, reduces the residual between the predicted and the actual parameters after the update. The second term, named \textit{reconstruction loss}, enforces that the actual parameters are a fixed-point of the iteration. During training, the perturbed parameters in the contraction loss are sampled from a Gaussian distribution with zero mean, and the variance is varied by sampling from a standard uniform distribution.
PUNet is also based on a transformer encoder, similar to NeuralSPMe. The causal mask is not applied to PUNet to allow the network to utilize the full sequence, and the average pooling is applied to the output of the transformer encoder to obtain the updated parameters from the output sequence. To allow PUNet to extract features for parameter update from all of the available inputs, the perturbed inputs $\tilde{\textbf{U}}_{i}$ are concatenated into a single sequence and provided to the PUNet as $\mathop{\Vert}\limits_{i=1}^{M}\tilde{\textbf{U}}_{i}$.
Figure~\ref{fig:ParameterEstimator} shows the architecture of PUNet.

\subsection{Parameter Identification via Fixed-Point Iteration}
\label{Proposed algorithm}
After obtaining NeuralSPMe $\bm{\Phi}$ and PUNet $\bm{\Psi}$ from training, the fixed-point iteration for parameter identification is performed as summarized in Algorithm~\ref{alg:proposed_algorithm}. The iteration consists of two steps: the forward step and the update step. In the forward step, lithium concentration and terminal voltage are evaluated by NeuralSPMe $\bm{\Phi}$. In the update step, the parameters are updated using PUNet $\bm{\Psi}$. The iteration continues until the RMSE between the evaluated and the target voltage becomes smaller than a predefined threshold.

\begin{algorithm}[H]
\caption{Neural acceleration $\bm{\Phi}$ and inverse surrogate $\bm{\Psi}$ for fixed-point battery parameter identification}
\label{alg:proposed_algorithm}
\begin{algorithmic}
\STATE Initialize $\bm{\lambda}^{0}, \textbf{x}_{i}^{0}, \textbf{y}_{i}^{0}, V_{i}^{0}$
\FOR{$k=0$ to $K$}
\IF{$\max\left(\text{RMSE}(V^{\text{ref}}_{i}-V^{k}_{i})\right)<\delta$}
\STATE \textbf{break}
\ENDIF
\STATE \textbf{do in parallel for $i=1,\dots,\text{M}$}
\STATE \quad $\textbf{y}_{i}^{k}=\bm{\Phi}(\textbf{x}_{i}^{k},\bm{\lambda}^{k},I_{i})$
\STATE \quad $V_{i}^{k}=h(\text{H}\textbf{y}_{i}^{k},\bm{\lambda}^{k},I_{i})$
\STATE \quad $\textbf{U}_{i}^{k}=[V_{i}^{k},\textbf{y}_{i}^{k},\bm{\lambda}^{k},V_{i}^{\text{ref}},I_{i}]$
\STATE \textbf{end for}
\STATE $\bm{\lambda}^{k+1}=\bm{\Psi}([\textbf{U}_{1}^{k}||\cdots||\textbf{U}_{M}^{k}])$
\ENDFOR
\end{algorithmic}
\end{algorithm}

\section{Data Acquisition and Preprocessing}
\label{Data Acquisition and Preprocessing}
\subsection{Acquisition of Base Parameters}
\label{Acquisition of Base Parameters}
To generate realistic synthetic data, we first determine the base parameters which represent a fresh cell. An high-capacity pouch cell for EV application is used in this work. First, the electrode thickness, particle radius, open-circuit potential of each electrode are measured. Then other parameters such as porosity, diffusion coefficients, cyclable lithium capacity are identified by minimizing the voltage error between the measured and simulated discharge curves. The discharge curves are obtained under CC discharge at \SI{25}{\degreeCelsius} at six different rates: 1/20~C, 1/10~C, 1/3~C, 1/2~C, 1~C, and 2~C. The simulation is performed using SPMe implemented in PyBaMM~\cite{sulzer2021python}, and the parameters are identified by minimizing the RMSE between the measured and simulated discharge curves. The covariance matrix adaptation evolution strategy (CMA-ES)~\cite{6790628} implemented in Optuna~\cite{optuna_2019} is used for the optimization. Note that this procedure is for obtaining realistic baseline parameters for the synthetic data generation described in~\ref{Synthetic Data Generation}.

\subsection{Synthetic Data Generation}
\label{Synthetic Data Generation}
After the reference parameters are determined, degraded versions of the parameters are sampled to generate the synthetic data for training and validation of NeuralSPMe and PUNet. The sampled parameters are filtered to yield a uniform distribution of state-of-health (SoH) in the range from 0.7 to 1.05, where SoH is defined as the ratio of the 1/3 C discharge capacity and the nominal capacity. The SoH range is quantized into 7 bins with a width of 0.05, and each bin contains 500 parameter sets. The range of each parameter except $c_{\text{s,p}}^{\text{max}}$ and $c_{\text{s,n}}^{\text{max}}$ are set to 50\% to 150\% of the values found in~\cite{chen2020development}, where the ranges of $c_{\text{s,p}}^{\text{max}}$ and $c_{\text{s,n}}^{\text{max}}$ are set to 75\% to 125\%.
The sampled parameters and their ranges are summarized in Table~\ref{table:sampled_params}.
\begin{table}[h]
    \centering
    \begin{tabular}{c c c}
        \hline
        \textbf{Parameter} & \textbf{Minimum} & \textbf{Maximum} \\
        \hline
        $\varepsilon_{\text{p}}$ & 0.137 & 0.400 \\
        $\varepsilon_{\text{n}}$ & 0.193 & 0.570 \\
        $R_{\text{p}}$ & \SI{2.98}{\micro\meter} & \SI{8.63}{\micro\meter} \\
        $R_{\text{n}}$ & \SI{7.72}{\micro\meter} & \SI{22.9}{\micro\meter} \\
        $c_{\text{s,p}}^{\text{max}}$ & \SI{4.17e4}{\mole\per\cubic\meter} & \SI{6.82e4}{\mole\per\cubic\meter} \\
        $c_{\text{s,n}}^{\text{max}}$ & \SI{2.92e4}{\mole\per\cubic\meter} & \SI{4.83e4}{\mole\per\cubic\meter} \\
        $D_{\text{s,p}}$ & \SI{2.97e-14}{\square\meter\per\second} & \SI{8.64e-14}{\square\meter\per\second} \\
        $D_{\text{s,n}}$ & \SI{4.31e-14}{\square\meter\per\second} & \SI{1.24e-13}{\square\meter\per\second} \\
        $Q_{\text{Li}}$ & \SI{65.4}{\ampere\hour} & \SI{100}{\ampere\hour} \\
        \hline
    \end{tabular}
    \caption{Sampled parameters $\mathring{\bm{\lambda}}$ and ranges.}
    \label{table:sampled_params}
\end{table}

\begin{figure}[t]
\begin{center}
\includegraphics[width=\linewidth]{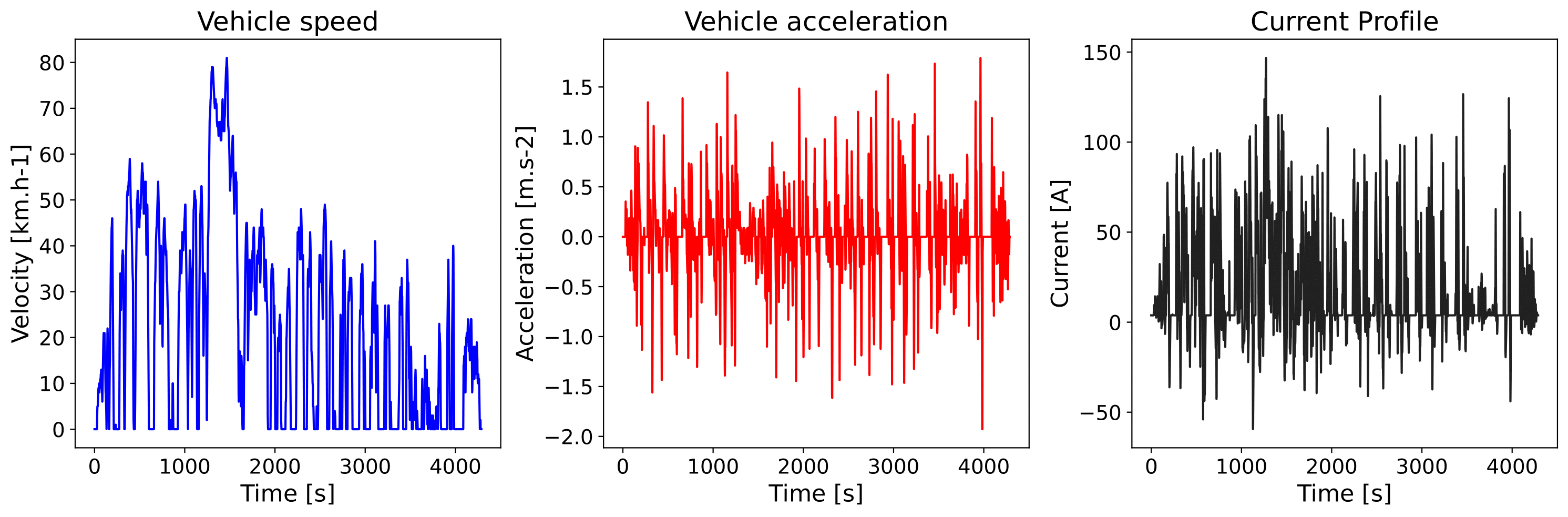}
\caption{An example of driving data and the generated current sequence.}    
\label{fig:driving_data}
\end{center}
\end{figure}

For each sampled parameter set, 10 simulations are performed with different initial SoC and current sequences. The current sequences are generated from three months of real-world driving data, which includes velocity, acceleration, and time stamps measured by GPS. The method for converting the driving data into current sequences is adopted from~\cite{plett2015battery}. An example of the driving data and the generated current sequence is illustrated in Figure~\ref{fig:driving_data}.

From the simulation results, the following data are obtained: terminal voltage, current, lithium concentration at the surface of the particles, average lithium concentration, and the average of the square root of lithium concentration in the positive electrolyte, along with the parameters to be identified. The obtained concentration values $\textbf{c}$ are transformed into the label, i.e., the normalized concentration values $\textbf{y}$ as in~(\ref{eq:NSPMe_output_conc}).
In addition, the parameter values are normalized to have zero mean and unit variance, the terminal voltage is scaled to the range of 0 to 1 (with 0 and 1 corresponding to the lower and upper cutoff voltages, respectively), and the current values are divided by 100.

The data are split into training and validation datasets, by selecting 50 parameter sets from each SoH bin for the validation dataset. The number of samples and parameter sets for training and validation datasets are summarized in Table~\ref{table:Data_summary}.
\begin{table}[t]
    \centering
    \begin{tabular}{c c c}
        \hline
        \textbf{Dataset} & \textbf{Training} & \textbf{Validation} \\
        \hline
        Number of samples & 31500 & 3500 \\
        Number of parameter sets & 3150 & 350 \\
        \hline
    \end{tabular}
    \caption{Summary of training and validation datasets.}
    \label{table:Data_summary}
\end{table}

\section{Results}
\label{Results}
\subsection{NeuralSPMe}
The accuracy of NeuralSPMe is evaluated on the validation dataset. To demonstrate the importance of the physics-embedded structure of NeuralSPMe, a pure data-driven model is trained as a baseline, which is a vanilla transformer (VT) encoder that directly predicts the terminal voltage from the current and parameters. Two different sizes of NeuralSPMe and VT models are trained. The hyperparameters of the small NeuralSPMe and the large NeuralSPMe are summarized in Table~\ref{table:NSPMe_hyperparams}. The RMSE of terminal voltage is used as a performance criterion.
\begin{table}[h]
    \centering
    \begin{tabular}{c c c}
        \hline
        \textbf{Hyperparameter} & \textbf{Small} & \textbf{Large} \\
        \hline
        Number of transformer encoder layers & 1 & 4 \\
        Number of attention heads            & 1 & 4 \\
        Embedding dimension of transformer & 8 & 96 \\
        Feedforward dimension of transformer & 16 & 192 \\
        \hline
        Number of trainable weights & 836 & 310372 \\
        \hline
    \end{tabular}
    \caption{Hyperparameters of small NeuralSPMe and large NeuralSPMe.}
    \label{table:NSPMe_hyperparams}
\end{table}

\begin{figure*}[t]
    \begin{center}
        \begin{subfigure}[t]{0.4\linewidth}
            \centering
            \includegraphics[width=\linewidth]{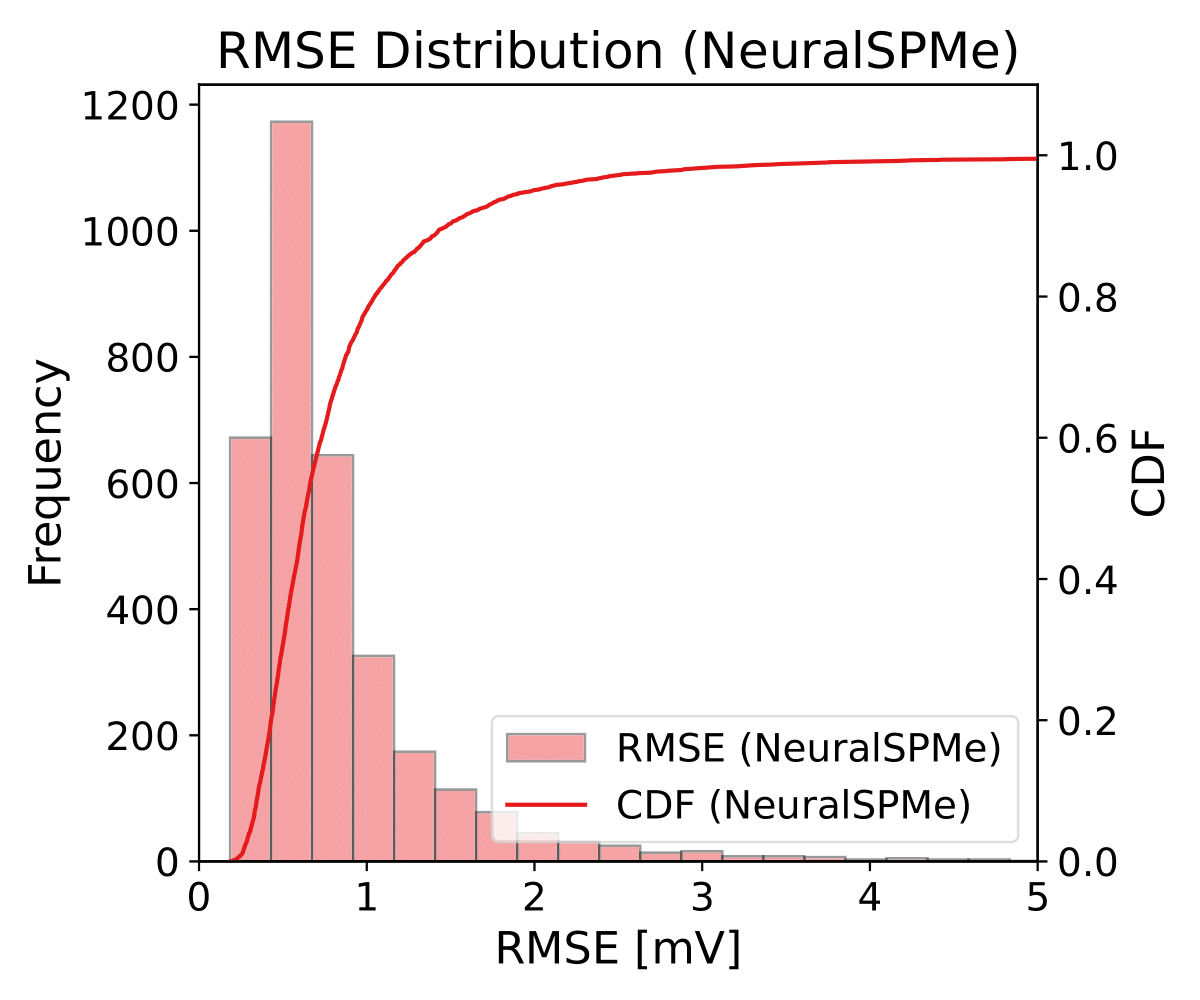}
            \caption{Voltage RMSE histogram of large NeuralSPMe.}
            \label{fig:RMSE_histogram_nspme}
        \end{subfigure}
        \hspace{0.05\linewidth}
        \begin{subfigure}[t]{0.4\linewidth}
            \centering
            \includegraphics[width=\linewidth]{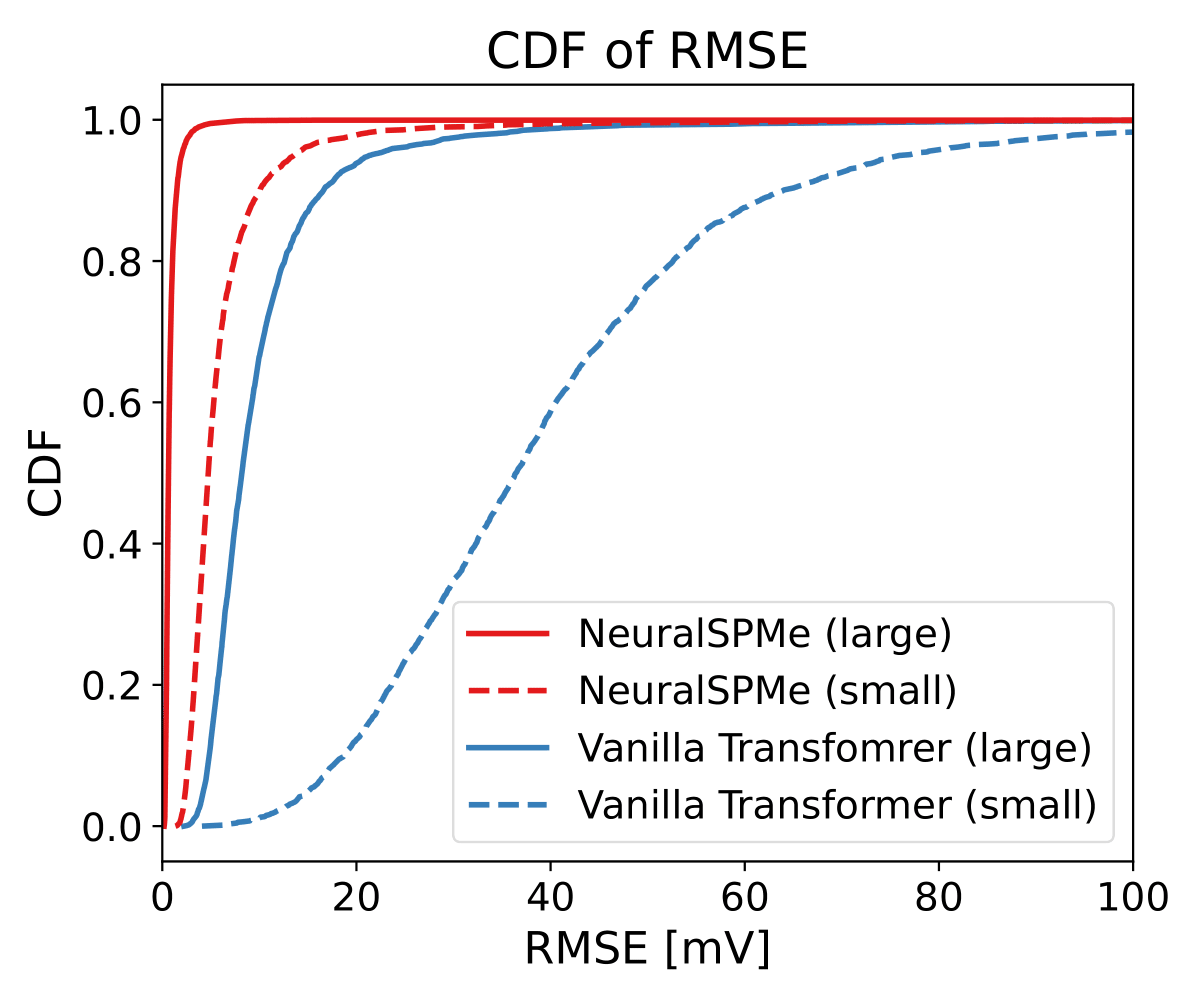}
            \caption{CDF of voltage RMSE of NeuralSPMe and VT models.}
            \label{fig:RMSE_cdf}
        \end{subfigure}
        \caption{Histogram and CDF of voltage RMSE of NeuralSPMe and VT models: (a) large NeuralSPMe, (b) all models.}
        \label{fig:RMSE_histogram}
    \end{center}
\end{figure*}

Our main model, i.e., the large NeuralSPMe shows the average RMSE of \SI{0.84}{\milli\volt} and the 90-th percentile RMSE of \SI{1.47}{\milli\volt} while the VT model shows \SI{10.23}{\milli\volt} and \SI{16.57}{\milli\volt}. The error between the large NeuralSPMe and SPMe is negligible compared to the error between SPMe and experimental measurements, which is around \SI{10}{\milli\volt}~\cite{dangwal2021parameter,shi2022physics,zhang2022beyond,ha2024cobrapro}. This result suggests that the large NeuralSPMe can replace SPMe without loss of accuracy in parameter identification.
Interestingly, even the small NeuralSPMe achieves the average RMSE of \SI{6.13}{\milli \volt} and outperforms the large VT model with only 370 times fewer trainable weights. This gap demonstrates that encoding of domain physics is much more beneficial rather than relying on large model capacity. Figure~\ref{fig:RMSE_histogram_nspme} presents the histogram and cumulative distribution function (CDF) of the RMSE for the large NeuralSPMe, and Figure~\ref{fig:RMSE_cdf} illustrates CDF of the RMSE for four models. As shown in Figure~\ref{fig:RMSE_cdf}, both the large and small NeuralSPMe models outperform the VT models.

There are several reasons why NeuralSPMe outperforms VT. First, NeuralSPMe takes the sequence of average stoichiometries calculated from a nonlinear solver and coulomb counting as input, which significantly reduces the learning burden of the model by providing good initial estimates of surface stoichiometries and relative positional information through entire sequence. Second, NeuralSPMe predicts the lithium concentrations rather than the terminal voltage so that the model can learn the physics described in SPMe. Third, the voltage expression in NeuralSPMe makes the voltage prediction much easier by providing the exact relationship between the lithium concentration, the parameters, and terminal voltage.

\begin{figure}[t]
    \begin{center}
        \begin{subfigure}[t]{\linewidth}
            \centering
            \includegraphics[width=\linewidth]{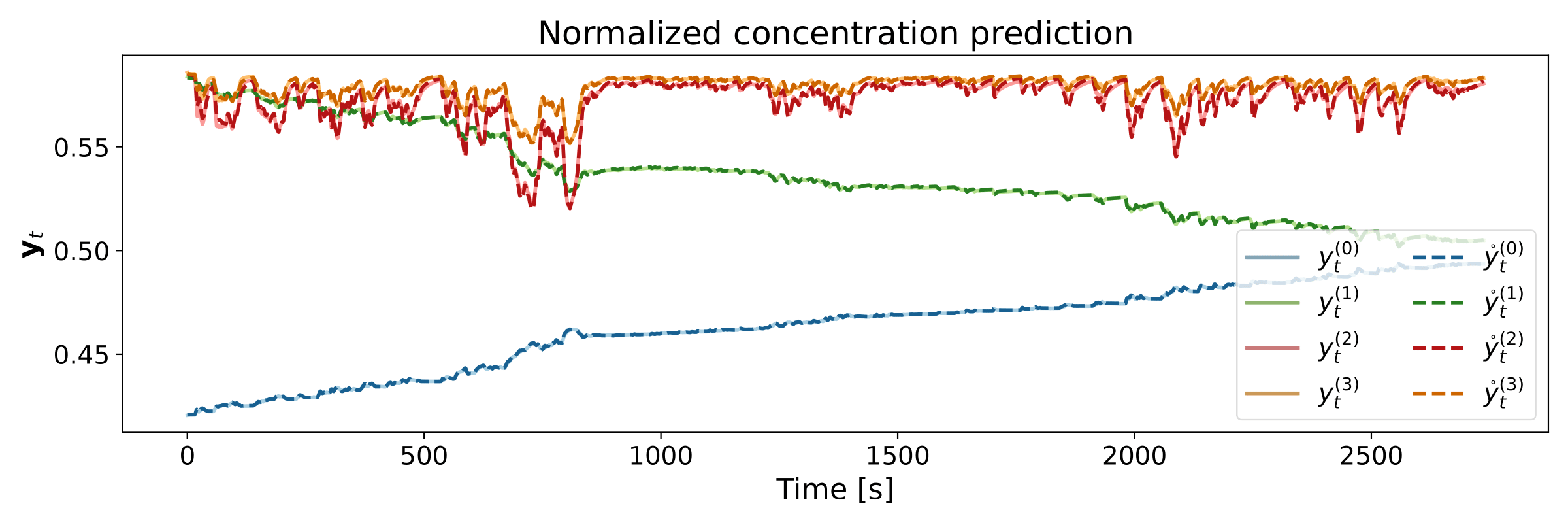}
            \caption{Normalized concentration prediction with large NeuralSPMe.}
            \label{fig:Concentration prediction}
        \end{subfigure}
        \begin{subfigure}[t]{\linewidth}
            \centering
            \includegraphics[width=\linewidth]{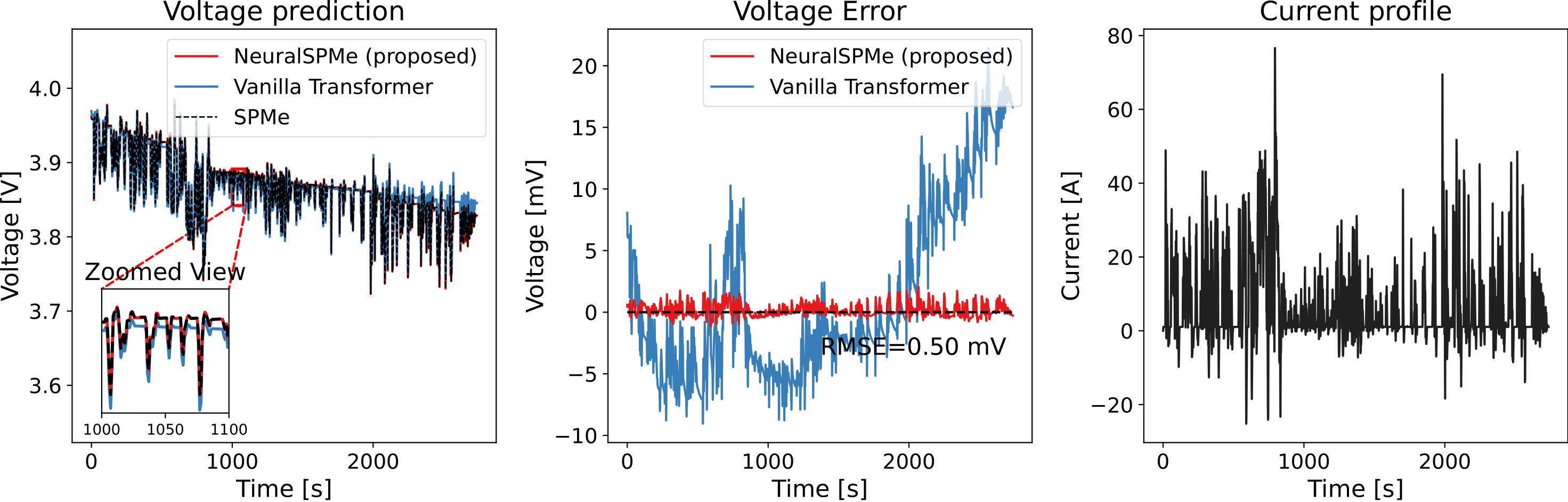}
            \caption{Voltage prediction with large NeuralSPMe and large VT model.}
            \label{fig:Voltage prediction}
        \end{subfigure}
        \caption{Prediction results of large NeuralSPMe: (a) Normalized concentration prediction and (b) Voltage prediction.}
        \label{fig:Combined prediction results}
    \end{center}
\end{figure}

Figure~\ref{fig:Concentration prediction} illustrates the normalized lithium concentration $\mathring{\textbf{y}}$ used as labels and the NeuralSPMe prediction $\textbf{y}$. As shown in the figure, NeuralSPMe can accurately evaluate the lithium concentrations at electrode surfaces and in the electrolyte. Figure~\ref{fig:Voltage prediction} shows the terminal voltage calculated from the predicted lithium concentration $\textbf{y}$, along with those from the large VT model and SPMe. The corresponding voltage error and current are also presented. 
The figure shows that NeuralSPMe accurately evaluates the peaks of the terminal voltage, resulting in minimal voltage error across a wide range of current values. The minimal error in predicted lithium concentration and terminal voltage demonstrates that NeuralSPMe has successfully learned the underlying physics of SPMe, highlighting the significant benefit of physics-embedded structure of NeuralSPMe.

\subsection{PUNet and the Proposed Algorithm}
Next we evaluate the proposed PUNet algorithm combined with the NeuralSPMe model. For comparison, parameter identification is performed on the validation dataset with three different methods: SPMe with CMA-ES~\cite{6790628}~(baseline), NeuralSPMe with CMA-ES, and NeuralSPMe with PUNet~(proposed). CMA-ES is chosen as the baseline because it provides robust performance with minimal hyperparameter tuning, unlike other metaheuristic algorithms such as PSO and GA. For CMA-ES, parameters are updated until the maximum voltage RMSE among 10 reference voltages becomes smaller than \SI{5}{\milli\volt}, and for PUNet, parameters are further updated until there is no improvement in the maximum voltage RMSE for three iterations.
The parameters are initialized with mean values of the training dataset. For SPMe with CMA-ES, 10 parameter sets are randomly selected from the validation dataset, and for other two methods, all 350 parameter sets in the validation dataset are used for evaluation. The experimental results are summarized in Table~\ref{table:PI_results_comparison} and details will be described below.
\begin{table}[h]
    \centering
    \begin{tabular}{l c c c}
        \hline
         & SPMe       & NeuralSPMe & NeuralSPMe \\
         & +CMA-ES     & +CMA-ES     & +PUNet      \\
                      & (baseline) &            & (proposed) \\
        \hline
        Mean time [s]       & 2786  & 56.19 & 1.32 \\
        Mean iterations     & 459.4 & 402.6 & 16.30 \\
        Speedup             & 1.00  & 49.60 times & 2113 times \\
        Sample efficiency   & -     & 1.00 & 24.70 times \\
        Parameter MAPE [\%] & -     & 26.7 & 1.76 \\
        \hline
    \end{tabular}
    \caption{Summary of average time and iterations taken for parameter identification with different methods.}
    \label{table:PI_results_comparison}
\end{table}

First we compare SPMe with CMA-ES and NeuralSPMe with CMA-ES to demonstrate the advantage of NeuralSPMe in terms of computation time. For 10 parameter sets, SPMe with CMA-ES takes 2786 seconds on average for a parameter identification, while NeuralSPMe with CMA-ES takes only 56.19 seconds on entire validation dataset. Thus, a speedup of 49.60~times is achieved by replacing SPMe with NeuralSPMe. Note that 10 simulations have to be performed to evaluate the objective function for each sampled parameter set, as described in~\ref{Synthetic Data Generation}. The 10 simulations are performed in parallel for both methods by multiprocessing on CPU and batched computation on GPU for SPMe and NeuralSPMe, respectively. 

Second, NeuralSPMe with CMA-ES and NeuralSPMe with PUNet are compared to measure the advantage of PUNet in terms of speedup and sample efficiency. For 350 parameter sets, CMA-ES takes 56.19 seconds on average as mentioned above, and PUNet takes only 1.32 seconds, achieving a speedup of 42.57~times compared to CMA-ES. This speedup comes from the superior sample efficiency of PUNet, which only takes 16.30 iterations on average while CMA-ES takes 402.6 iterations. This result demonstrates that PUNet is 24.70~times more sample efficient than CMA-ES. The proposed algorithm thus achieves 2113~times of speedup compared to the baseline, combining the advantages of NeuralSPMe and PUNet. Moreover, PUNet is much more accurate than CMA-ES in terms of mean absolute percentage error (MAPE) of the identified parameters. The MAPE of the identified parameters with PUNet is 1.76~\% on average, while CMA-ES shows 26.7~\%.

\begin{figure}[H]
\begin{center}
\includegraphics[width=0.8\columnwidth]{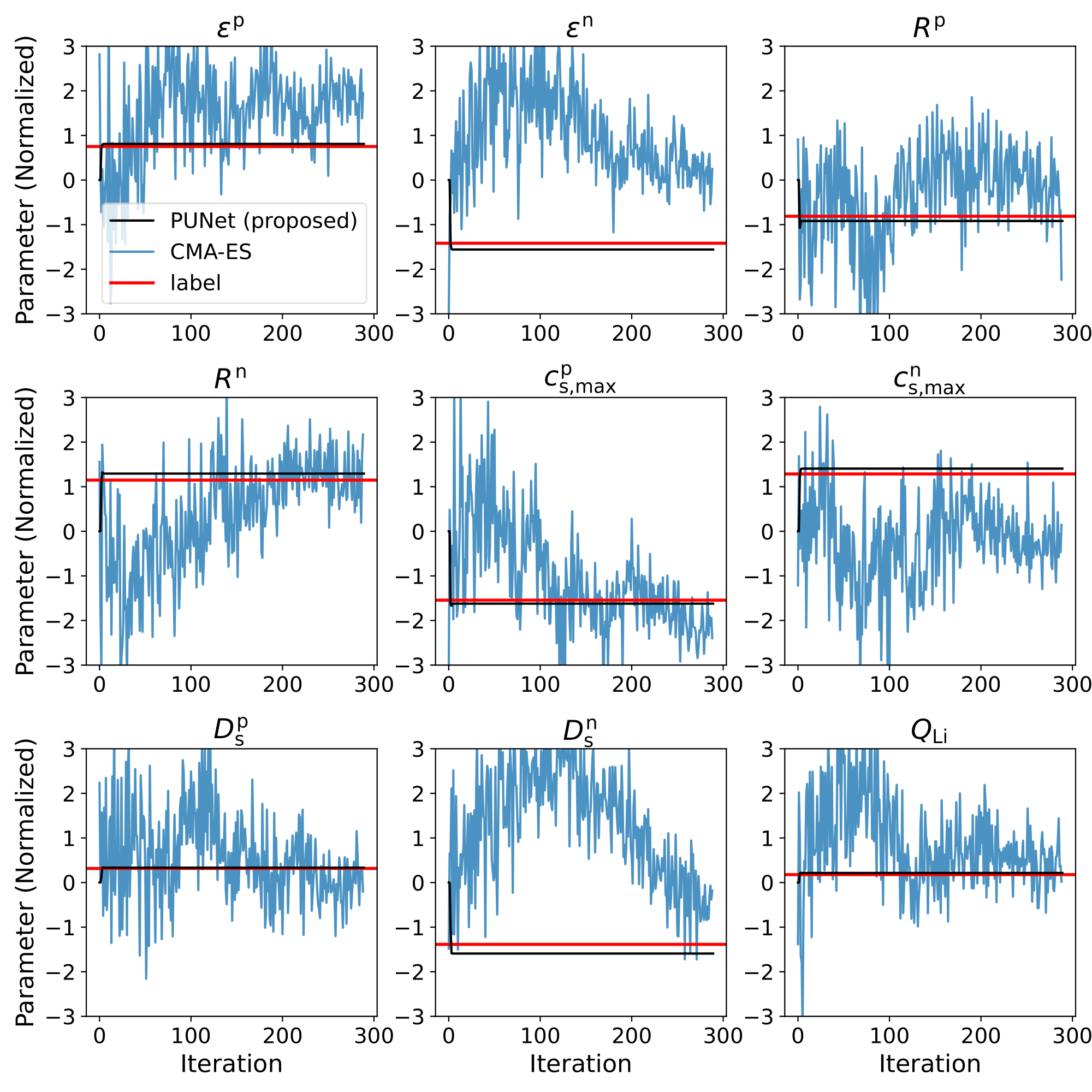}
\caption{Parameter update process of the proposed PUNet in comparison to CMA-ES.}
\label{fig:parameter_update_punet_cmaes}
\end{center}
\end{figure}

\begin{figure}[H]
    \begin{center}
        \includegraphics[width=0.8\columnwidth]{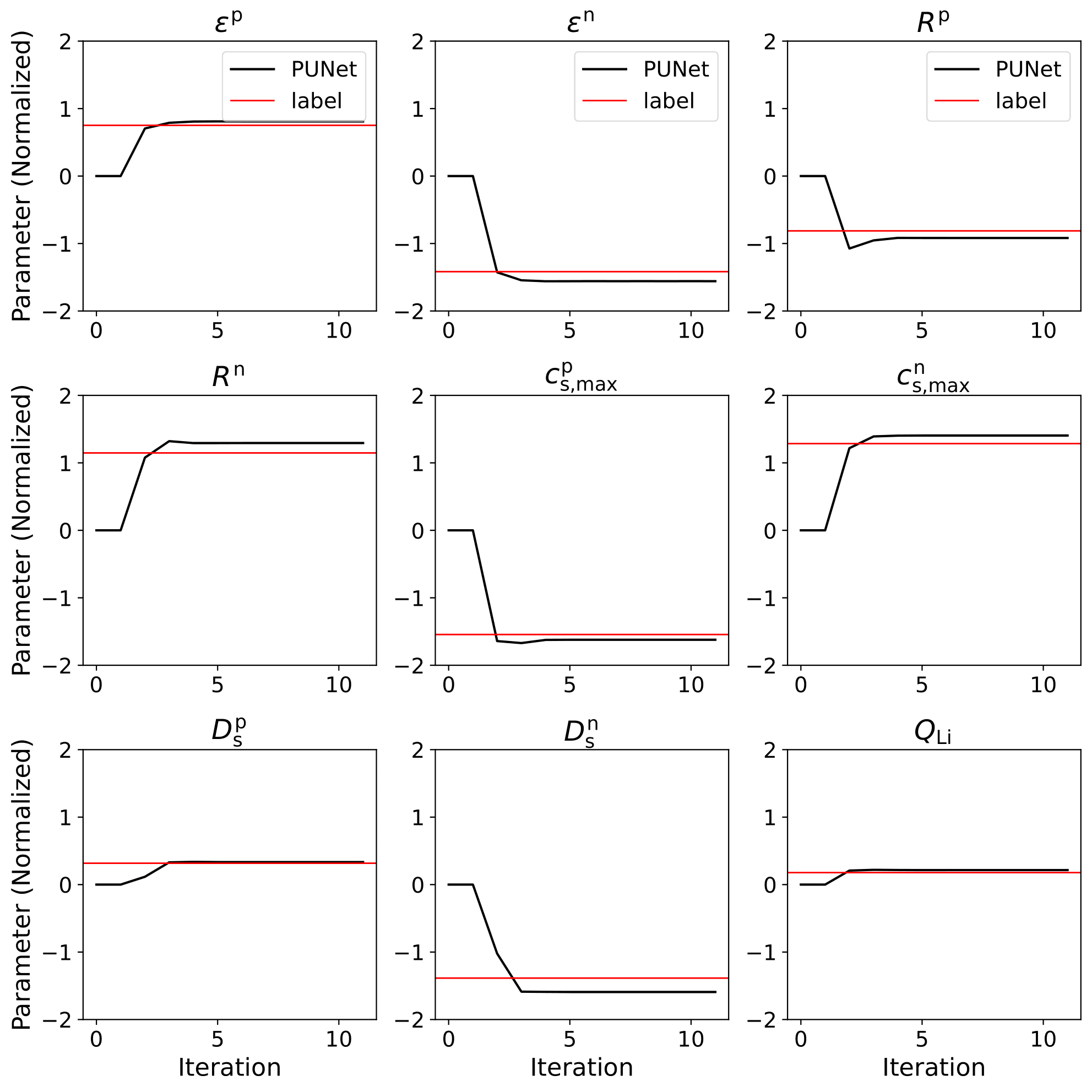}
        \caption{Fast convergence of PUNet (only a few iterations required).}
        \label{fig:parameter_update_punet}
    \end{center}
\end{figure}

\begin{figure}[H]
\begin{center}
    \includegraphics[width=0.8\columnwidth]{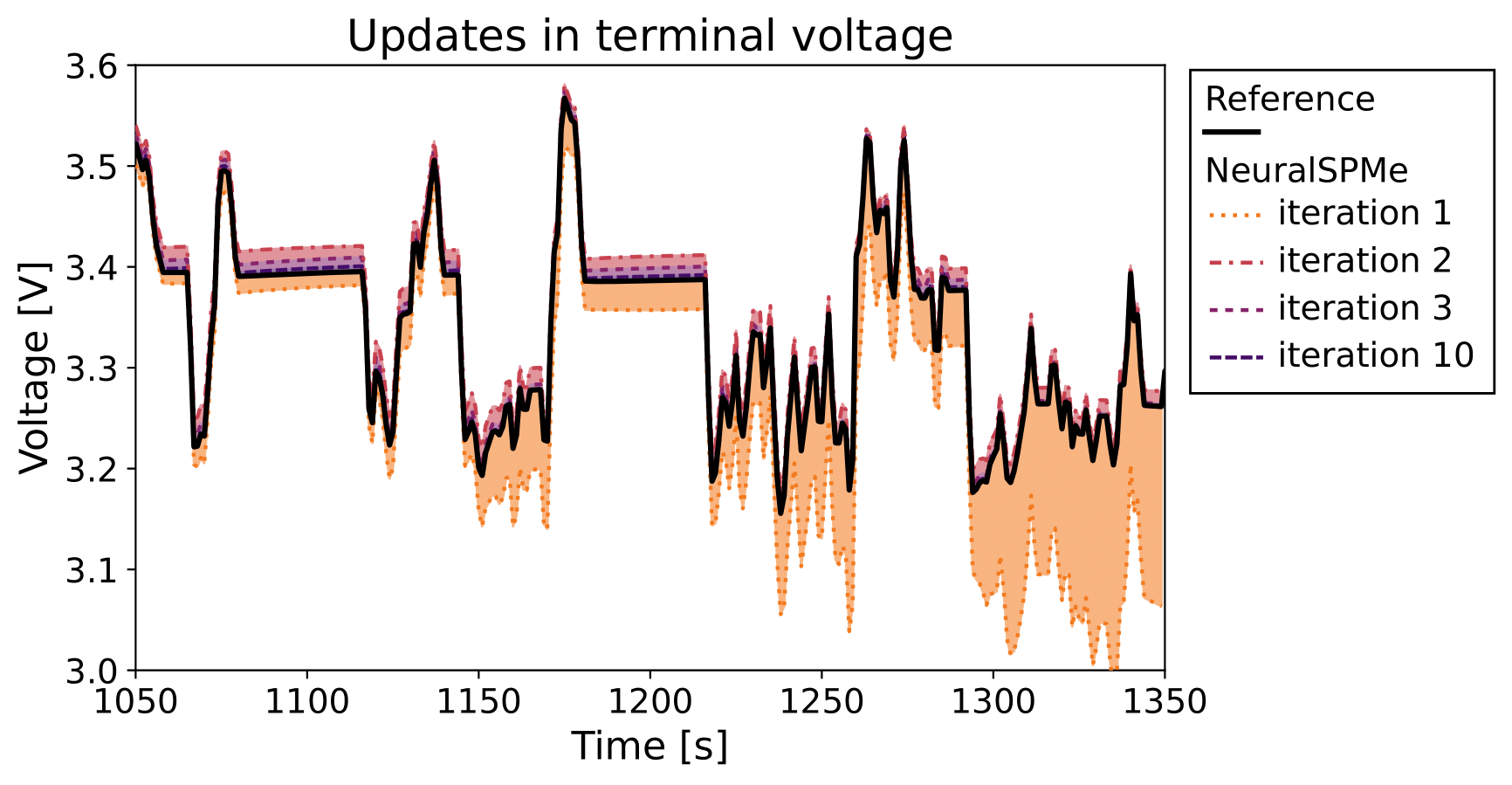}
    \caption{Terminal voltage evaluated by NeuralSPMe during parameter identification with PUNet.}
    \label{fig:voltage_updates}
\end{center}
\end{figure}

\begin{figure}[H]
\begin{center}
    \includegraphics[width=0.6\columnwidth]{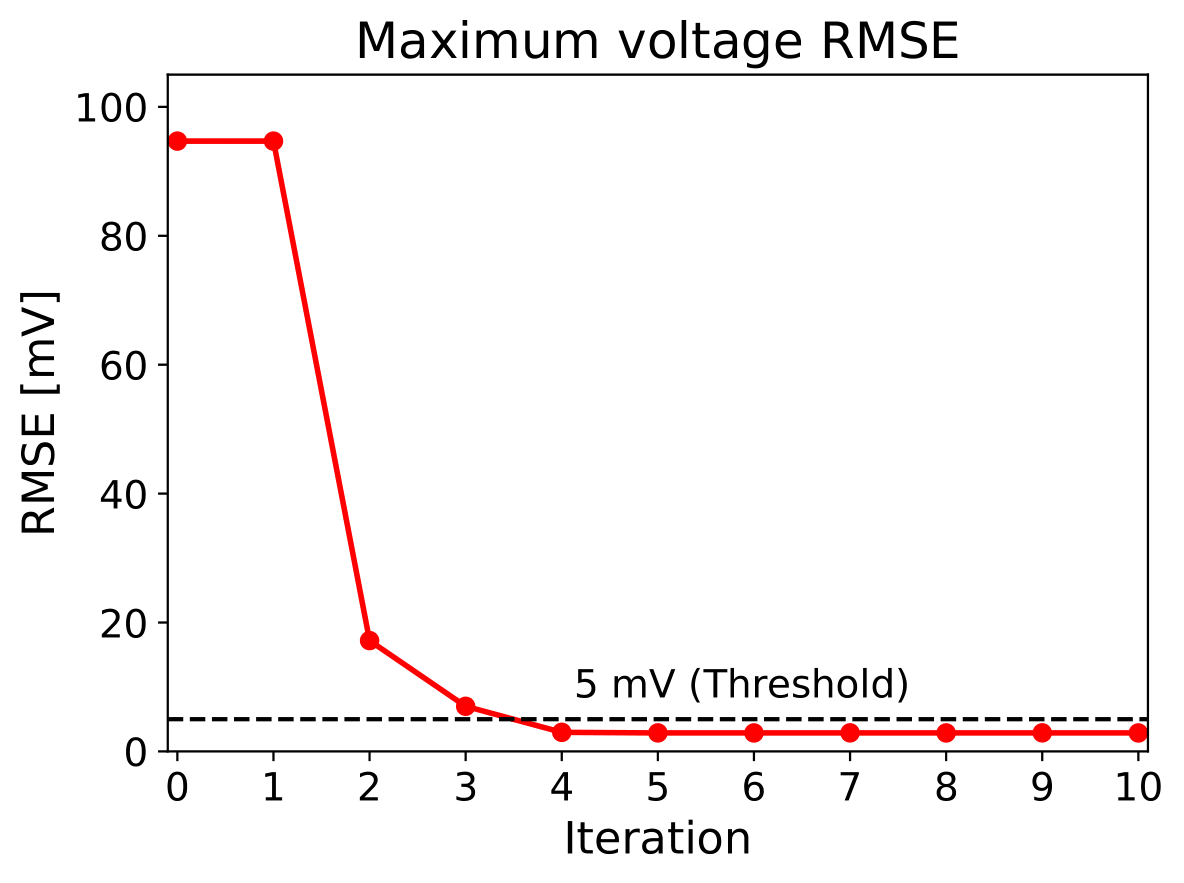}
    \caption{Decrease of maximum voltage RMSE out of 10 reference voltages during parameter identification with PUNet.}
    \label{fig:rmse_updates}
\end{center}
\end{figure}

\begin{figure}[t]
    \begin{center}
        \includegraphics[width=0.7\columnwidth]{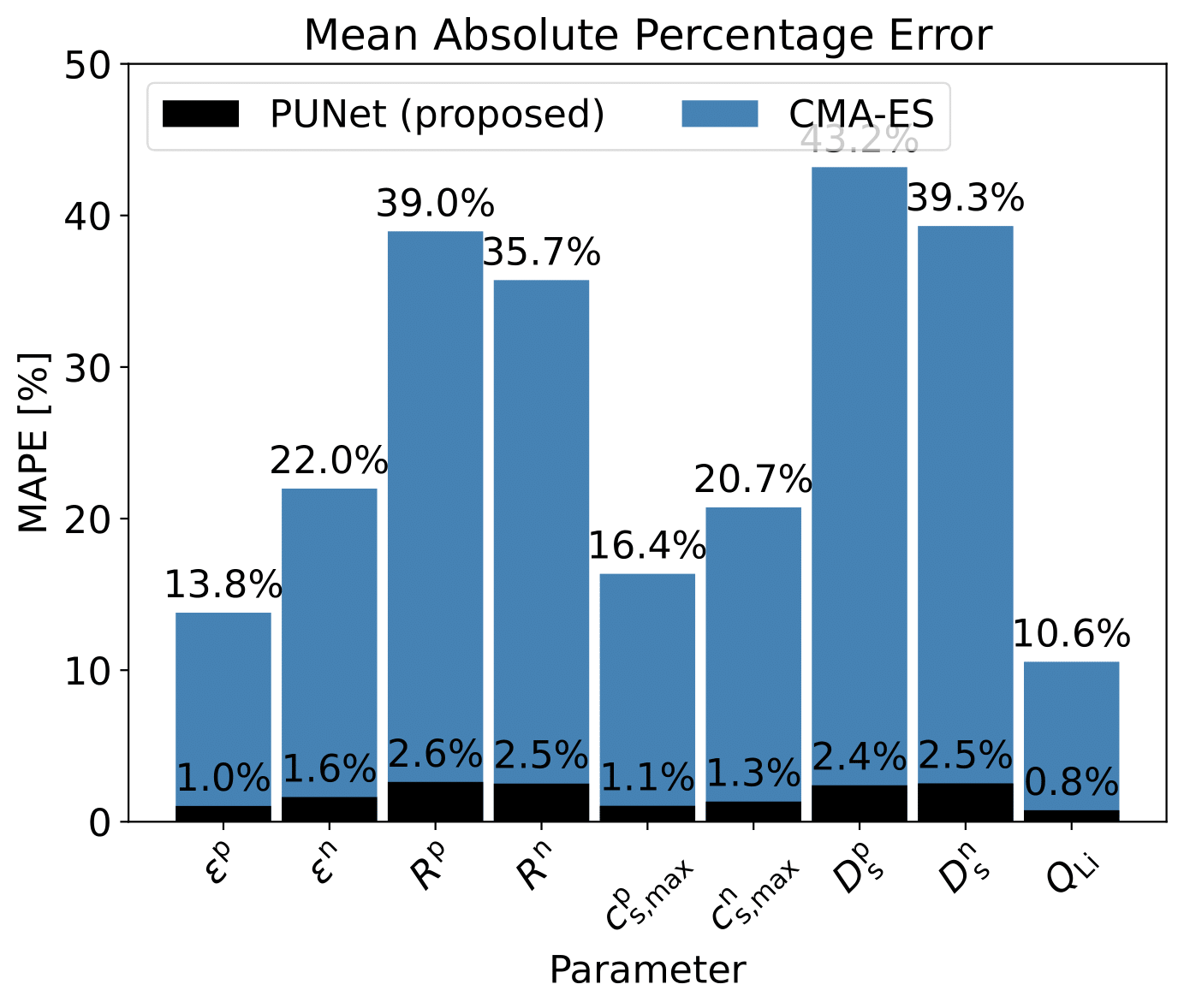}
        \caption{The MAPE of 350 parameter sets identified with PUNet and CMA-ES.}
        \label{fig:parameter mae}
    \end{center}
\end{figure}

Figure~\ref{fig:parameter_update_punet_cmaes} illustrates how the parameters are updated by PUNet and CMA-ES during parameter identification, and Figure~\ref{fig:parameter_update_punet} shows identical results for only PUNet. Surprisingly, the parameters updated by PUNet quickly approach to the actual values without any search in the parameter space as shown in Figure~\ref{fig:parameter_update_punet}. By contrast, Figure~\ref{fig:parameter_update_punet_cmaes} shows that CMA-ES performs hundreds of evaluations to find out which parameters are likely to have lower voltage error.
\noindent Figure~\ref{fig:voltage_updates} shows the terminal voltage evaluated by NeuralSPMe during parameter identification illustrated in Figure~\ref{fig:parameter_update_punet}. As shown in the figure, the evaluated voltage is quickly getting closer to the reference voltage as iteration proceeds. The RMSEs of evaluated voltages are presented in Figure~\ref{fig:rmse_updates}. The figure shows that the RMSE is reduced to below \SI{5}{\milli\volt} only at the fourth iteration, and further reduced to \SI{2.46}{\milli\volt} at the final iteration.
\noindent Finally, Figure~\ref{fig:parameter mae} shows the MAPEs of the identified values for each parameter. As can be seen, PUNet achieves significantly lower MAPE than CMA-ES, e.g., more than 10 times for every parameter, even with much fewer iterations. 

One may wonder why PUNet works so well. The fast and accurate parameter updates of PUNet come from the way it is trained. Most optimization algorithms including CMA-ES only utilize the scalar objective value, i.e., the average voltage RMSE, to evaluate sampled parameter sets. On the other hand, PUNet takes the evaluated voltage and lithium concentration as sequence, along with the reference voltage and current sequence so that it can fully utilize the rich information in the evaluation of parameter sets. Moreover, since PUNet is trained with the parameter labels, which conventional optimization algorithms cannot utilize, PUNet is more accurate in terms of errors in the identified parameters.
\FloatBarrier

\section{Conclusions}
\label{Conclusions}
This paper proposed a deep learning-based framework for a fast parameter identification of lithium-ion batteries under dynamic operating conditions. The proposed framework consists of NeuralSPMe, a neural surrogate model for SPMe, and PUNet, an inverse surrogate model for parameter update. The proposed framework performs a fixed-point iteration to identify the parameters with NeuralSPMe and PUNet.
NeuralSPMe is a physics-embedded transformer encoder that evaluates the lithium concentrations and terminal voltage. By combining a numerical solver and coulomb counting with transformer architecture, NeuralSPMe evaluated terminal voltage 49.6~times faster than SPMe with only \SI{0.84}{\milli\volt} of an average RMSE, while the vanilla transformer model showed \SI{10.23}{\milli\volt}. 
PUNet further accelerated the parameter identification by conducting parameter updates with its superior sample efficiency. We compared the proposed PUNet with a metaheuristic optimization algorithm, CMA-ES, and demonstrated that PUNet shows 24.7~times higher sample efficiency. Furthermore, PUNet is much more accurate than CMA-ES because it is trained with the parameter labels, which conventional optimization algorithms cannot utilize. 
By leveraging NeuralSPMe and PUNet, the proposed framework performed parameter identification in only 1.32 seconds on average, which is 2113~times faster than SPMe with CMA-ES and achieved 10-fold lower MAPE than CMA-ES for every parameter. In our future works for real-world applications, practical issues such as measurement noise and thermal effects can be addressed on top of the developed rapid parameter identification.

\bibliographystyle{elsarticle-num}
\bibliography{cas-refs}

\end{document}